\documentclass[10pt]{wlscirep}

\AtBeginDocument{%
  \providecommand\BibTeX{{%
    \normalfont B\kern-0.5em{\scshape i\kern-0.25em b}\kern-0.8em\TeX}}}
    
\usepackage[utf8]{inputenc}
\usepackage[T1]{fontenc}
\usepackage{lineno}
\usepackage{multirow}
\usepackage{array}
\usepackage{subfigure}
\usepackage{algorithm}
\usepackage{algorithmicx}
\usepackage{algpseudocode}
\usepackage{color}
\usepackage{bm}
\usepackage{tabularx}
\usepackage{colortbl}  
\usepackage{xcolor} 
\usepackage{hhline}
\usepackage{newtxtext}
\usepackage{newtxmath}
\usepackage[flushleft]{threeparttable}

\title{Sculpting Molecules in Text-3D Space: A Flexible Substructure Aware Framework for Text-Oriented Molecular Optimization}

\author[1,$\dag$]{Kaiwei Zhang}
\author[2,$\dag$]{Yange Lin}
\author[3]{Guangcheng Wu}
\author[2]{Yuxiang Ren}
\author[2]{Xuecang Zhang}
\author[2,4]{Bo Wang}
\author[1]{Xiaoyu Zhang}
\author[2,5,*]{Weitao Du}

\affil[1]{Institute of Information Engineering, Chinese Academy of Sciences, Beijing 100085, China}
\affil[2]{Huawei Technologies}
\affil[3]{Department of Chemistry, The University of Hong Kong, Hong Kong SAR 999077, China}
\affil[4]{School of Chemistry and Chemical Engineering, Harbin Institute of Technology, Harbin 150001, China}
\affil[5]{Academy of Mathematics and Systems Science, Chinese Academy of Sciences}
\affil[*]{Corresponding author:  Weitao Du (duweitao@mail.ustc.edu.cn)}
\affil[$\dag$]{These authors contributed equally to this work.}

\begin{abstract}
The integration of deep learning, particularly AI-Generated Content, with high-quality data derived from ab initio calculations has emerged as a promising avenue for transforming the landscape of scientific research. However, the challenge of designing molecular drugs or materials that incorporate multi-modality prior knowledge remains a critical and complex undertaking.
Specifically, achieving a practical molecular design necessitates not only meeting the diversity requirements but also addressing structural and textural constraints with various symmetries outlined by domain experts. 
In this article, we present an innovative approach to tackle this inverse design problem by formulating it as a multi-modality guidance optimization task. Our proposed solution involves a textural-structure alignment symmetric diffusion framework for the implementation of molecular optimization tasks, namely 3DToMolo. 3DToMolo aims to harmonize diverse modalities including textual description features and graph structural features, aligning them seamlessly to produce molecular structures adhere to specified symmetric structural and textural constraints by experts in the field. 
Experimental trials across three guidance optimization settings have shown a superior hit optimization performance compared to state-of-the-art methodologies. Moreover, 3DToMolo demonstrates the capability to discover potential novel molecules, incorporating specified target substructures, without the need for prior knowledge.
This work not only holds general significance for the advancement of deep learning methodologies but also paves the way for a transformative shift in molecular design strategies. 3DToMolo creates opportunities for a more nuanced and effective exploration of the vast chemical space, opening new frontiers in the development of molecular entities with tailored properties and functionalities.
\end{abstract}
\begin{document}

\flushbottom

\maketitle

\section{Introduction}
In the realm of molecule optimization, a pivotal undertaking in drug discovery, and chemical engineering including catalyst and polymer material designs, the imperative lies in enhancing the desired properties of candidate molecules through strategic chemical modifications. This pivotal process revolves around the core objective of generating molecules that not only meet stringent structural, physical, and electrochemical criteria, but also retain essential structural features (beneficial for relatively straightforward and economic synthesis \cite{Chen2021}). At the center of this inverse design issue is the fact that the targeted properties are diverse, spanning the spectrum from qualitative to quantitative aspects. These encompass properties dependent on electronic structures to overarching global descriptions, intricately tied to the two-dimensional (2D) bond topology of the molecule. A concomitant challenge emerges from the multifaceted nature of the goals, necessitating the manipulation of scales and modalities within the molecule. This spans the gamut from fine-tuning atom types and topological structure of atoms to orchestrating alterations in the three-dimensional (3D) conformer structures, reflecting the varied and nuanced nature of the optimization objectives. As a result, traditional solutions rely on the knowledge and expertise of medicinal chemists, often executed through fragment-based screening or synthesis \cite{gerry2018chemical, hoffer2018integrated, deSouzaNeto2020InSS}. However, such approaches are inherently limited by their lack of scalability and automation.

In recent years, the landscape of computational lead generation has witnessed the emergence of \textit{in silico} methodologies. These methodologies prominently feature deep learning techniques such as latent-space-based generation and Monte-Carlo tree searching (MCTS) algorithms, which trade explicit mechanistic interpretability to model more complex biological relationships learned directly from data such as SMILES (Simplified Molecular Input Line Entry System) \cite{kusner2017grammar, gomez2018automatic, sanchez2018inverse, segler2018generating} and two-dimensional molecular graphs \cite{duvenaud2015convolutional, liu2019n, liu2023moleculestm, zeng2022deep}. The ensuing consequence is the flourishing advancement in the field of molecular discovery, driven by the intricate challenges inherent in identifying novel compounds endowed with specific and desired properties. Within this expansive domain, one prominent line of research focuses on generative models, such as variational autoencoders (VAEs) \cite{gomez2018automatic, kusner2017grammar, nakata2018molecular, simonovsky2018graphvae} and generative adversarial networks (GANs) \cite{goodfellow2020generative, prykhodko2019novo, gomez2018chemgan, de2018molgan}, which leverage deep learning \cite{krishnan2021novo, arus2019randomized, bagal2021molgpt, mahmood2021masked, gupta2018generative, li2021structure} techniques to generate novel molecules. These models have demonstrated promising results in generating diverse and chemically valid molecules. By formulating the molecule optimization problem as a sequence-to-sequence or graph-to-graph translation problem,  \cite{He2021,Hoffman2022} also utilizes molecular autoencoders as the backbone model for purely 2D molecule optimization. Another approach entails the employment of Reinforcement Learning (RL) algorithms to iteratively optimize molecular structures guided by predefined objectives. RL-based methods \cite{atance, popova2018deep, olivecrona2017molecular,putin2018reinforcement,you2018graph, segler2018planning} have shown potential in optimizing drug-like properties and exploring chemical space efficiently. 

However, a notable gap persists in the utilization of traditional encoder-decoder-based \textit{de novo} molecule generation methods for molecule optimization tasks. Lead optimization \cite{jorgensen2009efficient} focuses on improving the properties of existing lead compounds, leveraging experimental data and medicinal chemistry expertise to systematically refine molecular structures. This approach tends to retain the major scaffold of molecules for yielding drug candidates with better-defined pharmacological profiles and higher likelihoods of success in clinical trials. Unlike models such as normalizing flows, GANs, and VAEs that generate molecules in a zero-shot manner from informationless noise, effective automatic molecule optimization demands the learning of the distribution differences between molecules before and after the optimization, aligning with preferred properties. MoleculeSTM \cite{liu2023moleculestm} addresses this challenge by introducing a latent optimization block that guides property-directed transformations through vector movements in the latent space. Since this occurs in the latent space rather than the real 3D molecular space, such approaches grapple with diversity collapsing issues, potentially leading to the loss of crucial molecular structure information. On the other hand, implicit searching-based methods, such as reinforcement learning and MCTS, necessitate expert-designed optimization paths. These paths are instrumental in training the reward function, ensuring it aligns with fixed properties, and formulating policies for molecular modifications. In practice, this entails identifying disconnection optimization sites, such as optimal side chains, at each step. A learned policy network then selects the best actions from a pre-fixed set of valid molecule modifications. However, this approach may suffer from inflexibility, as the predefined optimization path data and the modification set may not capture the diverse and nuanced possibilities inherent in molecule optimization.

In general, it is highly desirable to develop a methodology that is purposefully tailored for optimizing both 2D (atom types and chemical bond topology) and conformer structure (3D) aspects of molecules. Simultaneously, this methodology should exhibit compatibility with a broad spectrum of complex goals, facilitating multi-goal guidance optimization. Capitalizing on the remarkable capabilities exhibited by large language models (LLMs), there is a natural inclination to explore the feasibility of consolidating property and structure descriptions into a unified text format. Then, we are able to leverage the prowess of LLMs to extract a unified representation from such textual amalgamation.  In pursuit of this, we advocate the training of a joint molecule diffusion model designed to capture the fine-grained distributions of 2D+3D molecule structures. The crux of an ideal molecule optimization lies in achieving alignment within the representation spaces of both the text side and the molecule structure side. To this end, we introduce the \underline{3D}-based \underline{T}ext-\underline{o}riented \underline{Mol}ecular \underline{O}ptimization (3DToMolo), wherein this specific cross-modality alignment is realized through contrastive training. This involves training the representation pair obtained from a lightweight LLM and an (SE(3)) equivariant graph transformer specifically tailored for molecules.
The intermediary steps introduced during the forward diffusion process play a crucial role as a medium connecting the initial molecules with those possessing target properties. In contrast to generative approaches of sampling from white noise, the intermediate molecule representations retain essential structural information from the original molecules. Moreover, control over text descriptions is meticulously exerted at each step during the subsequent backward optimization process. Beyond the flexibility inherent in optimizing entire regions of molecules, 3DToMolo showcases its prowess in two practical scenarios where substructures are preserved. In these instances, specific three-dimensional structures are pre-fixed, and optimization exclusively occurs within the remaining inpainting areas, highlighting the versatility and effectiveness of 3DToMolo.

\section{Results}
\subsection{Definition of text - structural optimization}
 
Natural-language texts provide a cohesive framework for articulating intricate details regarding the structural and property characteristics of molecules. We follow the approach presented by MoleculeSTM \cite{liu2023moleculestm} for optimizing structures of molecules, guided by textual prompts. These prompts may encompass qualitative and quantitative descriptions, addressing single or multiple goals. Nevertheless, a notable limitation of the latent space optimization approach proposed in MoleculeSTM lies in its lack of 3D structure encoding. It is imperative to recognize that 3D conformer structures of molecules determine their 2D chemical bond relations, and contribute significantly to their chemical and physical properties. Consequently, successful optimization of molecules or well-known scaffolds with precisely tuned properties requires the integration of 3D structures. 

\textbf{Task definition.\ } Given a molecule or molecular fragment $M_0$ with known 2D and 3D structures, molecule optimization aims to modify atom types, 3D positions and associated bond relations \cite{o2011open} to produce another molecule $M_y$. This transformation is guided by the prompt-text $y$, ensuring that  $M_y$ aligns better with the given text than the original molecule $M_0$. 

To establish a connection between the original molecule structure and the optimized molecule, we introduce a series of noised states $M_t$. Coarsening fine-grained details, $M_t$ is a blurred version preserving essential semantic information. With a well-selected time horizon $T$, $M_T$ serves as a common representation bridging $M_0$ and $M_y$. Utilizing diffusion-based generative models, known for their efficacy in generating molecule graphs \cite{jo2022score} and 3D conformers \cite{liu2023group}, we propose that parameterizing and controlling the denoising process, which reverses $M_T$ to $M_0$, provides a flexible and grounded method for optimizing molecules.

Suppose $M_t$ is generated by a Markov chain defined as:
\begin{equation} \label{eq: nosing}
dM_t = f(M_t,t) dt + g(t)\cdot dW_t,    
\end{equation}
where $W_t$ denotes Brownian motion, and $f$ and $g$ are smooth functions depending on the current molecules and time $t$. Let $p_t(M_t)$ be the marginal distribution of the noised molecule $M_t$. A $\theta$ parameterized $SE(3)$-equivariant graph transformer $S_{\theta}$ is employed to learn the gradient of the log-likelihood $p_t(M_t)$: $\nabla \log p_t(M_t)$. The optimizing process with prompt $y$ follows the formula:
\begin{equation} \label{eq: optimizing process}
dM = [f(M,t) - g^2(t)\cdot \nabla \log p_t(M, y)]dt + g(t)\cdot dW_t,  
\end{equation}
where $\nabla \log p_t(M, y) = \nabla \log p_t(M) +  \nabla \log p_t(y | M)$. Fitting the conditional probability $p_t(y | M)$ involves using the latent molecular embedding extracted from another graph transformer, trained independently with $S_{\theta}$ (by pairing with the text embedding of the prompt $y$).  We will outline the overall workflow of 3DToMolo in the next section.

\begin{figure}[htbp]
	\centering
	\begin{minipage}[c]{\textwidth}
		\centering
		\includegraphics[width=0.8\textwidth]{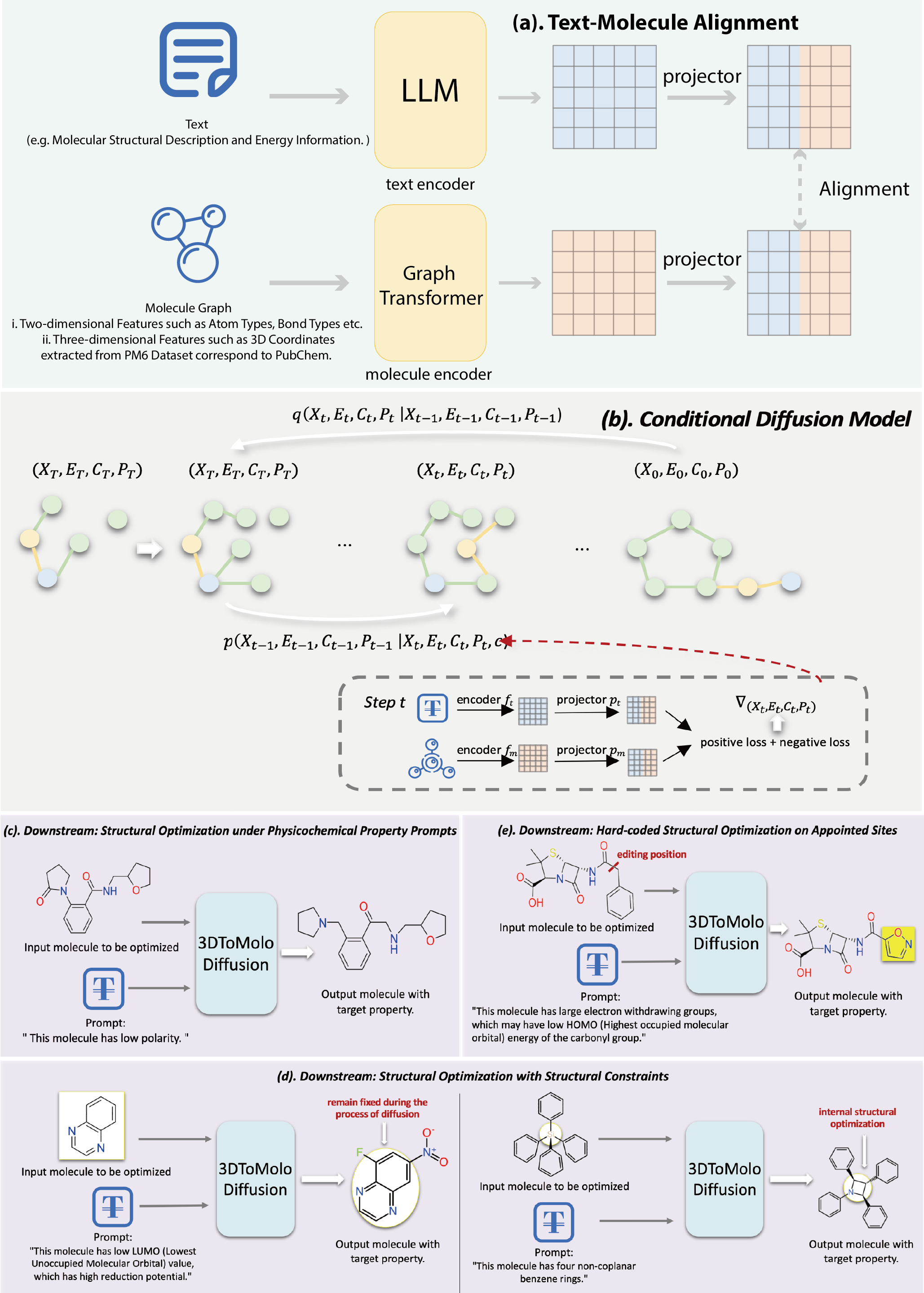}
        \vspace{-1mm}
	\end{minipage}
 \captionsetup{font={small}}
	\caption{Overview of 3DToMolo. \textbf{(a).} The alignment of textual description and chemical structures of molecules, which is realized through contrastive learning of the two latent representations: molecule structure encoding with its paired text embedding. \textbf{(b).} Conditional diffusion model. In order to maintain molecule optimization in alignment with the prompt, conditional diffusion model incorporates text prompts at each step during the subsequent backward optimization process. \textbf{(c).} The zero-shot prompt-driven molecule optimization task involves modifying the input molecule in response to a given text prompt related to physicochemical properties. 3DToMolo necessitates the overall optimization of both 2D and 3D features of molecules, ensuring a balanced alignment with the input molecule and the text prompt which is achieved by the conditional diffusion model, shown as (b). \textbf{(d).} Molecule optimization under structural constraints. This task further enhances the similarity to the input molecule by retaining essential structural features. \textbf{(e).} Molecule optimization under appointed sites. Given the precise position within the input molecule, 3DToMolo aims to optimize molecule by offering strategies for atoms and the bonds connected with the site.}
	\label{fig:Arch}
\vspace{-60pt}
\end{figure}

\subsection{Development of a text-structural diffusion model} 
3DToMolo unfolds in two phases: pretraining and the subsequent application of pretrained models to three types of downstream optimization tasks, as illustrated in Figure \ref{fig:Arch}. During the pretraining phase, two key objectives are pursued. First, the alignment of textual descriptions and chemical structures is undertaken. Second, an unconditional 2D+3D molecular generation model is initiated. For both objectives, we employ an encoder-decoder-based equivariant graph transformer that takes the 2D molecular graph and the 3D coordinates of each atom as input. However, for the first goal, we exclusively utilize the encoder component to extract the latent representation of molecules. This decoupled workflow enables the utilization of extensive structural data lacking accompanying text descriptions for training $S$. This aspect is crucial for the generation of diverse optimized structures.

On the prompt-text embedding side, we leverage the widely acclaimed large language model, LLAMA \cite{touvron2023llama}, as the text encoder, tapping into its ability to capture nuanced semantic representations from textual descriptions. Then, the alignment is achieved through contrastive learning of the two latent representations: the molecule structure encoding and its paired text embedding. As a possible extension, the text-structure alignment can be independently fine-tuned for domain-specific texts, e,g., materials \cite{rubungo2023llmprop}. To validate the effectiveness of our learned molecule latent embedding, we conduct tests on retrieval and property prediction tasks. The experimental results are provided in Table \ref{tab:prediction results}. In line with prior research on molecule pretraining \cite{hu2019strategies, liu2023moleculestm}, we adopt the MoleculeNet benchmark \cite{wu2018moleculenet}, which encompasses eight single-modal binary classification datasets aimed at evaluating the efficacy of pretrained molecule representation approaches. 
We consider nine pretraining-based methods as baselines: AttrMask \cite{hu2019strategies}, ContextPred \cite{hu2019strategies}, InfoGraph \cite{Sun2019InfoGraphUA}, MolCLR \cite{Wang2021MolCLRMC}, GraphMVP \cite{liu2019n}, MoleculeSTM \cite{liu2023moleculestm}, GEM \cite{fang2022geometry}, and MGSSL (BFS/DFS) \cite{zhang2021motif}.
We adopt the area under the receiver operating characteristic curve (ROC-AUC) \cite{bradley1997use} as the evaluation metric. As delineated in Table \ref{tab:prediction results}, our observations reveal that methods based on pretraining markedly enhance overall classification accuracy compared to randomly initialized counterparts. Additionally, 3DToMolo demonstrates superior performance on five out of eight tasks, while achieving comparable results to the leading baselines in the remaining three tasks. Since 3DToMolo lies in its ability to leverage pretrained chemical structure representations that incorporate external domain knowledge, which potentially provides a beneficial implicit bias for property prediction tasks. The key hyperparameters of molecule encoder are layers of Graph Transformer $5$, learning rate $\{1e-4, 1e-5\}$, hidden states dimension (X: 256, E: 128, pos: 64).

\begin{table}[!htbp]

  \caption{Downstream prediction results conducted on eight binary classification datasets sourced from MoleculeNet. The mark '-' represents the randomly initialized method.}
  \centering
  \label{tab:prediction results}
  \begin{tabular}{ccccccccc}
    \hhline{=========}
    methods & BBBP & Tox21 & Bace & ToxCast & Sider & ClinTox & MUV & HIV\\
    \cline{1-9}
    - & 68.67 & 73.09 & 77.89 & 63.40 & 59.03 & 68.97 & 70.94 & 77.28 \\
    \cline{1-9}
    AttrMask & 67.79 & 75.00 & 80.28 & 63.57 & 58.05 & 75.44 & 73.76 & 75.44 \\
    \cline{1-9}
    ContextPred & 63.13 & 74.29 & 78.75 & 61.58 & 60.26 & 80.34 & 71.36 & 70.67 \\
    \cline{1-9}
    InfoGraph & 64.84 & 76.24 & 77.64 & 62.68 & 59.15 & 76.51 & 71.97 & 70.20 \\
    \cline{1-9}
    MolCLR & 67.79 & 75.55 & 71.14 & 64.58 & 58.66 & 84.22 & 72.76 & 75.88 \\ 
    \cline{1-9}
    GraphMVP & 68.11 & 77.06 & 80.48 & 65.11 & 60.64 & 84.46 & 74.38 & 77.74\\
    \cline{1-9}
    MoleculeSTM & 69.98 & 76.91 & 80.77 & 65.06 & 60.96 & 92.53 & 73.40 & 76.93\\
    \cline{1-9}
    GEM & 72.40 & \textbf{78.10} & \textbf{85.60} & \textbf{69.20} & 67.20 & 90.10 & \textbf{81.70} & 80.60 \\
    \cline{1-9}
    MGSSL (BFS) & 69.70 & 76.50 & 79.10 & 64.10 & 61.80 & 80.70 & 78.70 & 78.80 \\
    \cline{1-9}
    MGSSL (DFS) & 70.50 & 76.40 & 79.70 & 63.80 & 60.50 & 79.70 & 78.10 & 79.50 \\
    \cline{1-9}
    3DToMolo & \textbf{72.50} & 77.16 & 85.53 & 67.11 & \textbf{67.88} & \textbf{93.18} & 77.65 & \textbf{80.66}\\
    \hhline{=========}
    \end{tabular}

\end{table}

Additionally, we pretrain an unconditional diffusion model designed as the backbone to capture the vast and complex data distribution and generate new structures within the chosen chemical space. The diffusion model samples Gaussian noise and undergoes iterative denoising, resembling standard diffusion sampling. 
The validity of our pretrained generative models is verified through standard chemical validity tests, as detailed in Supplementary Table S6.  
The introduction of the datasets we used is provided in the method section.

By integrating the alignment objective into the denoising process (with the noising step $t$ iteratively tuned to retain the similarity ratio between $M_y$ and $M_0$), as detailed in the method section, the model is capable of optimizing molecules with desired properties in a seamless, end-to-end manner. Notably, 3DToMolo is zero-shot, signifying that throughout the optimization process. We refrain from introducing any feedback for multi-stage correction, and the denoising process is executed in a single run.
We propose three types of downstream tasks in the following sections and systematically verify the robust effectiveness of 3DToMolo for text-structural optimization. All downstream tasks conducted are based on zero-shot optimization, which refers to a scenario where a model is required to optimize or adapt to new tasks or data without having been explicitly trained on those specific tasks or data beforehand. For instance, in molecular optimization, a model might be able to modify a molecule to achieve a desired biological activity, even if it has never been trained on molecules with that specific task, by leveraging its understanding of chemical space learned during pretraining. Zero-shot optimization is particularly valuable because it allows for the exploration of new chemical spaces or the design of molecules with novel properties, without requiring a model to be retrained or fine-tuned on every new task.

\subsection{Flexible molecule optimization under Physicochemical Property Prompts}
According to the degree of human knowledge involved in the molecule optimization process, we may classify them into two categories:

1. \textbf{Flexible optimization:} This category encompasses processes that do not explicitly specify the sites for optimizing atoms and the bonds connected with them.

2. \textbf{Hard-coded optimization:} In contrast, this category involves processes that precisely indicate the locations for optimization or substructures to be retained, providing hard constraints regarding the targeted atoms and their geometry.

Details of both optimization algorithms are available in Supplementary Section S4. Here, we prioritize the first category, wherein the prompt exerts a global influence on the entire molecular structure without specifically identifying optimization sites. Formally, given a molecule $M_0$ to be optimized, we adopt the following pipeline:
$$M_0 \rightarrow M_T \xrightarrow{y} M_y, $$
where $y$ denotes the text-prompt. We deliberately choose a small value for T, ensuring that the Tanimoto similarity coefficient \cite{bajusz2015tanimoto} between $M_T$ and $M_0$ approaches unity. While our text prompts encompass constraints ranging from 2D structure considerations (e.g., the number of hydrogen bond donors or acceptors) to properties determined by 3D structure (e.g., polarity), we primarily focus on analyzing how 3DToMolo effectively utilizes 3D structure information to enhance the alignment of the optimized molecule with the given text prompt. It is important to note that, in addition to energies directly calculated from 3D electronic configurations, we are equally intrigued by properties that, while validated through the generated SMILES, might demonstrate indirect connections with 3D structures throughout the optimization process. Our goal is to investigate whether our optimization process, which involves 3D structures, takes advantage of such properties. Consequently, we have designed multi-objective prompts, such as "soluble in water and having high polarity," to assess whether the 3D structure constraint, specifically the requirement for polarity, aids in guiding our optimization process through the vast chemical space.

In order to provide a reasonable and comprehensive evaluation of the molecule optimization ability of 3DToMolo, we first benchmarked representative state-of-art machine learning baselines, including:
\begin{itemize}
    \item \textbf{MoleculeSTM:} A multi-modal model, which enhances molecule representation learning through the integration of textual descriptions. 

    \item \textbf{GPT-3.5:} With its immense language processing capabilities and a broad understanding of chemistry concepts, has the potential to revolutionize molecule optimization.
    \item \textbf{Galactica:} A versatile scientific language model, extensively trained on a vast repository of scientific text and data.
\end{itemize}
Note that while the training data for Large Language Model (LLM)-based models encompasses significantly more scientific texts than domain-specific models, such as ours, it is limited in its ability to assimilate information from modalities other than textual, such as 3D structures. The effectiveness of models is assessed through a satisfactory hit ratio, indicating whether the output molecule generated by the model aligns with the conditions specified in the text prompt when given both a text prompt and a molecule for optimization.

\begin{table*}
  \caption{Results on 18 text prompts oriented towards diverse physicochemical objectives. The inputs consist of 200 molecules randomly sampled from ZINC, with the evaluation measured by the hit ratio (\%) of the property changes in each experiment. Best baseline results are
highlighted with underlined text. Best overall results are marked by *. Statistically significant improvement (t-test over 5 different dataset splits, p-value$<$ 0.05) is highlighted with bold text. }
    \centering
  \label{tab:overall performance}
  \begin{tabularx}{\textwidth}{m{0.5\textwidth}<{\raggedright}|m{0.15\textwidth}<{\centering}|X<{\centering}|X<{\centering}|X<{\centering}}
    \toprule
    \multirow{2}{*}{Prompts} & \multicolumn{3}{c|}{Baselines} & \multirow{2}{*}{3DToMolo} \\
    \cline{2-4}
     & MoleculeSTM & GPT-3.5 & Galactica & \\
    \midrule
    This molecule has low HOMO (Highest occupied molecular orbital) value, which is more stable. & 35.00 & \underline{37.50} & 00.00 & $\textbf{46.50}^*$ \\
    \cline{1-5}
    This molecule has high HOMO (Highest occupied molecular orbital) value, which is more reactive and susceptible to electron acceptance or participation in chemical reactions. & 28.00 & \underline{40.50} & 00.00 & $\textbf{58.50}^*$ \\
    \cline{1-5}
    This molecule has low LUMO (Lowest Unoccupied Molecular Orbital) value. & 09.50 & \underline{27.50} & 00.00 & $\textbf{30.00}^*$ \\
    \cline{1-5}
    This molecule has high LUMO (Lowest Unoccupied Molecular Orbital) value. & \underline{53.50} & 30.50 & 00.00 & $\textbf{88.00}^*$ \\
    \cline{1-5}
    This molecule has low HOMO-LUMO gap value, which has enhanced light absorption properties. The small energy difference allows the molecule to absorb photons in the visible or ultraviolet range, resulting in a higher likelihood of exhibiting color or being used as a dye or pigment. & 09.00 &  \underline{29.50} & 00.00 & $\textbf{87.50}^*$ \\
    \cline{1-5}
    This molecule has high HOMO-LUMO gap value, which is insulating or non-conductive. The large energy difference between the HOMO and LUMO orbitals makes it less likely for electrons to be excited across the gap, resulting in low electrical conductivity. & \underline{56.50} & 40.00 & 00.00 & $\textbf{89.00}^*$ \\
    \cline{1-5}
    This molecule has high polarity. & 41.50 & \underline{44.50} & 00.00 & $\textbf{58.00}^*$\\
    \cline{1-5}
    This molecule has low polarity. & \underline{45.00} & 35.50 & 00.00 & $\textbf{73.50}^*$ \\
    \cline{1-5}
    This molecule is soluble in water, which may have high polarity. & \underline{28.50} &  25.00 & 04.00 &  $\textbf{46.00}^*$ \\
    \cline{1-5}
    This molecule is insoluble in water, which may have low polarity.  & 52.00 & \underline{74.00} & 03.50 & $\textbf{81.00}^*$ \\
    \cline{1-5}
    This molecule is soluble in water. & $\underline{29.50}^*$  & 24.50  & 04.00 & 26.50 \\
    \cline{1-5}
    This molecule is insoluble in water. & 52.00 & \underline{66.50} & 03.50 & $\textbf{88.00}^*$ \\
    \cline{1-5}
    This molecule has high permeability. & \underline{34.50} & 23.00 & 05.50 & $\textbf{89.00}^*$\\ 
    \cline{1-5}
    This molecule has low permeability. & 24.50 & $\underline{39.00}^*$ & 00.50 & 21.00 \\
    \cline{1-5}
    This molecule has more hydrogen bond acceptors. & 07.50 & $\underline{20.00}^*$ & 00.00 & 10.50 \\
    \cline{1-5}
    This molecule has more hydrogen bond donors. & 05.00 & \underline{12.50} & 00.00 & $\textbf{34.50}^*$ \\
    \cline{1-5}
    This molecule is like a drug. & 42.00 & $\underline{59.00}^*$ & 00.00 & 41.50 \\
    \cline{1-5}
    This molecule is not like a drug. & \underline{39.00} & 31.00 & 00.00 & $\textbf{56.50}^*$ \\ 
    \cline{1-5}
    
    \bottomrule
  
\end{tabularx}

\end{table*}

Table \ref{tab:overall performance} summarizes the molecule optimization performance of 3DToMolo and existing approaches on a randomly selected subset of 200 molecules from the Zinc dataset \cite{Zinc}. To establish a zero-shot generalization stage for testing 3DToMolo, these 200 molecules were intentionally excluded from the model's pre-training data (although they may be present in other baseline models depending on their respective training datasets). We delve into 18 optimization tasks encompassing both 2D and 3D-related optimization. The tasks cover a wide range of energetic and structural properties of molecules (scientific background in Supplementary Section S2). It is evident that 3DToMolo consistently achieves exemplary hit ratios across the majority of the 18 tasks. This observation underscores the validity and benefits of incorporating 3D structures of molecules into the diffusion model and aligning chemical space with semantic space, thereby facilitating the exploration of output molecules satisfied with the desired properties. Incorporation of 3D structures also provides an additional navigation for exploring the accessible chemical space. Our optimization results diversify from different input molecules, different prompts, and different parallel runs (Supplementary Figure S6). This feature enables 3DToMolo to effectively improve the hit ratio by conducting multi-run optimization (Supplementary Table S4).


\begin{figure}[!htbp]
    \centering
    \includegraphics[width=\linewidth]{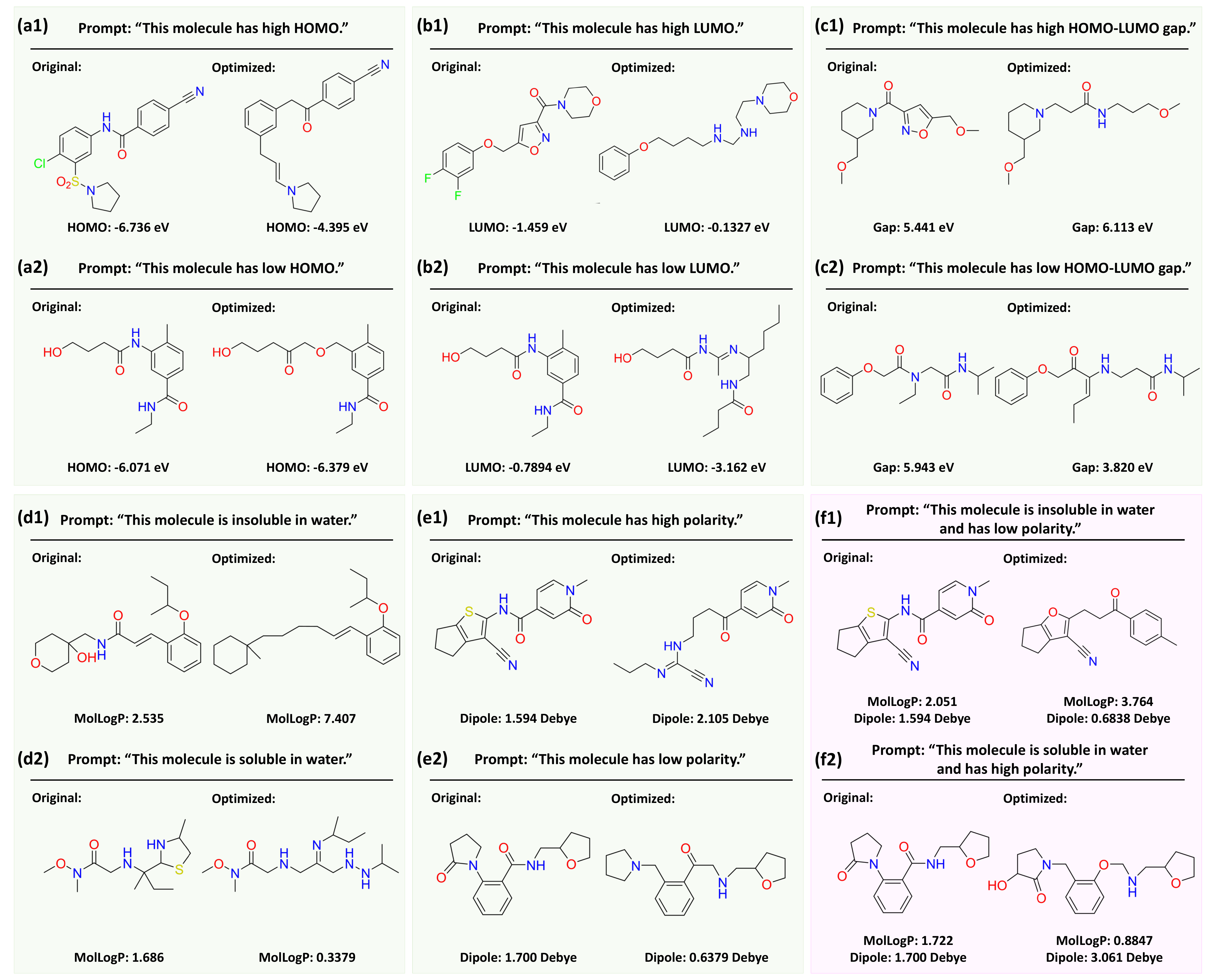}
    \caption{\label{fig:analysis} Exemplary prompt-driven molecule optimizations: \textbf{(a)} the highest occupied molecular orbital (HOMO) energy optimization, \textbf{(b)} the lowest unoccupied molecular orbital (LUMO) energy optimization, \textbf{(c)} the HOMO-LUMO energy gap optimization, \textbf{(d)} the water solubility optimization, \textbf{(e)} the polarity optimization, \textbf{(f)} the water solubility and polarity multi-objective optimization. Prompts shown in the figure are simplified and exact prompts used in experiments can be found in Table \ref{tab:overall performance}.}
\end{figure}

\begin{figure}[!htbp]
    \centering
    \begin{minipage}[c]{\textwidth}
	\centering
	\includegraphics[width=\textwidth]{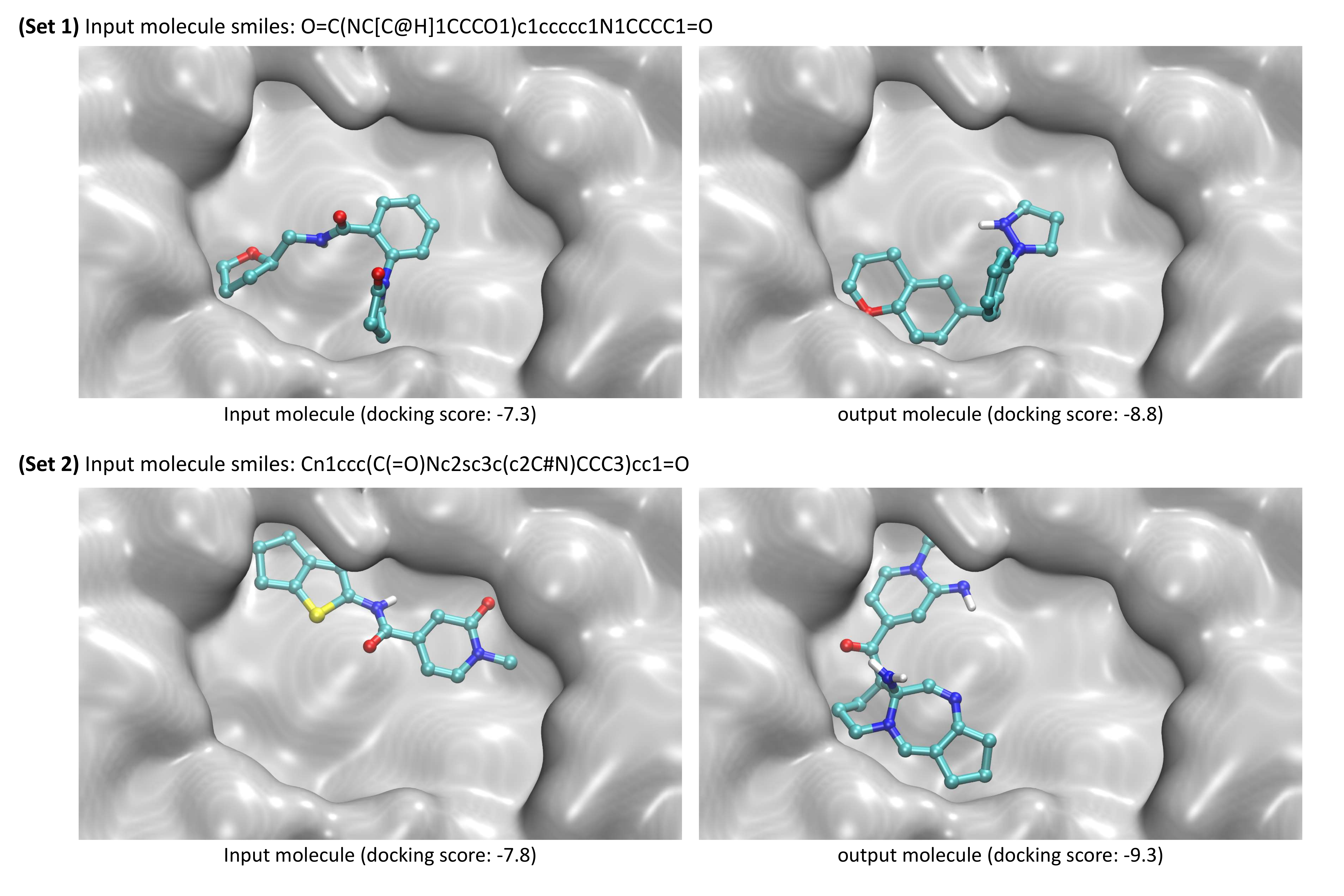}
    \end{minipage}
    \caption{The visualization of binding-affinity-based molecule optimization. The text prompt is from ChEMBL 1613777 ("This molecule is tested positive in an assay that are inhibitors and substrates of an enzyme protein. It uses molecular oxygen inserting one oxygen atom into a substrate, and reducing the second into a water molecule." \cite{mendez2019chembl}).   }
    \label{fig:docking}
\end{figure}

\textbf{Visual analysis on single-objective molecule optimizations.} We conduct a detailed visual analysis of disparities between original and optimized molecules, focusing on the single-objective tasks. Common modifications involve the addition, removal, and replacement of functional groups or molecular cores, with frequent occurrences of molecular skeleton rearrangements due to our ability to manipulate three-dimensional structures. For instance, atoms with high electronegativity have large electron affinities and thus can lower overall electron energy levels, whereas atoms with low electronegativity do the opposite. Therefore, to heighten electron energy levels in response to tasks like increasing the HOMO energy (Figure \ref{fig:analysis}(a1)) and the LUMO energy (Figure \ref{fig:analysis}(b1)), 3DToMolo removes highly electronegative atoms and functional groups in the input molecule, such as fluorine and chlorine atoms, as well as sulfone and isoxazole groups. Conversely, in tasks requiring the reduction of electron energy levels, electron-withdrawing functional groups or atoms with high electronegativity are introduced (Figure \ref{fig:analysis}(a2) and \ref{fig:analysis}(b2)). In Figure \ref{fig:analysis}(c1), the widening of the HOMO-LUMO gap is achieved by replacing the isoxazole group with a saturated chain. In contrast, Figure \ref{fig:analysis}(c2) showcases the narrowing of the HOMO-LUMO gap through the introduction of a double bond conjugated with the carbonyl group. This is because the introduction/removal of conjugated structure can result in denser/sparser electron energy levels and hence a wider/narrower HOMO-LUMO gap. Figure \ref{fig:analysis}(d1) and \ref{fig:analysis}(d2) illustrate that the addition and removal of hydrogen bond-forming groups, like hydroxyl groups and amines, modulate aqueous solubility by increasing and decreasing it, respectively. Concerning molecular polarity, optimizations such as changing a sulfur atom to a nitrogen atom increase bond polarity, enhancing overall polarity (Figure \ref{fig:analysis}(e1)), while the removal of a polar carbonyl group decreases polarity (Figure \ref{fig:analysis}(e2)). Additionally, we conduct binding-affinity-based molecule optimization. As shown in Figure \ref{fig:docking}, two sets of output molecules have lower docking scores, validating that the ligands generated by 3DToMolo could bind the receptor with higher affinity.

\textbf{Visual analysis on multi-objective molecule optimizations.} We further analyze the multi-objective molecule optimization. Water solubility and polarity are two positively correlated properties. Consequently, 3DToMolo turns the 2-oxo-1-pyrindinyl group into a benzene group, which reduces the solubility as well as the polarity of the input molecule (Figure \ref{fig:analysis}(f1)). In contrast, 3DToMolo adds a hydroxyl group to the input molecule when given the opposite prompt, which increases the solubility and the polarity (Figure \ref{fig:analysis}(f2)). More results on multi-objective optimization are presented in Supplementary Table S5. We observe that in the multi-objective task of improving both the solubility and the polarity, $46\%$ of the input molecules have been observed a solubility improvement after optimization, while $31.5\%$ have improvements in both properties, higher than the hit ratio of the single-objective solubility improvement task (Table \ref{tab:overall performance}). It hints that coupling with the polarity in the prompt helps us better tune the solubility. A possible reason could be that 3DToMolo tunes the polarity more flexibly, as discussed in the next paragraph.

\begin{figure}[!htbp]
    \centering
    \includegraphics[width=0.9\linewidth]{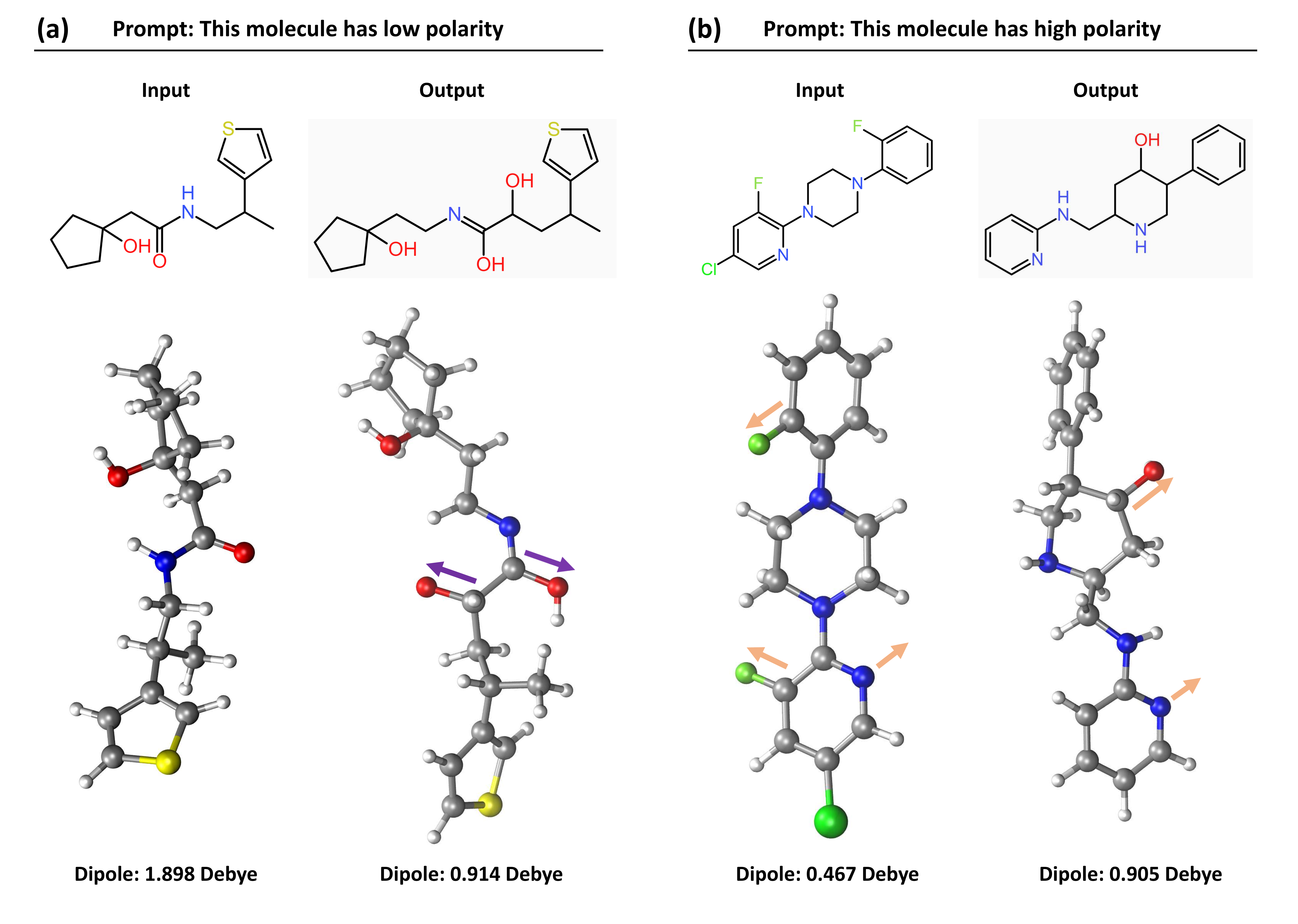}
    \caption{\label{fig:3d_polarity} Exemplary optimizations involving spatial information in the polarity-related prompts. Above are the 2D graphs of molecules and below are their corresponding 3D conformations. \textbf{(a)} Under the prompt "This molecule has low polarity", a hydroxyl group is added to the molecule, which neutralizes the dipole of the neighboring hydroxyl group due to the opposite alignment, as illustrated by the arrows. Consequently, the dipole moment of the molecule is reduced from 1.898 Debye to 0.914 Debye. \textbf{(b)} Under the prompt "This molecule has high polarity", the output molecule discards two C-F bonds which counteract the dipole of the pyridine ring and hence do not contribute much to the polarity. The removal of C-F bonds and the introduction of an aligned hydroxyl group raise the dipole moment of the molecule from 0.467 Debye to 0.905 Debye.}
\end{figure}

\textbf{Case study for 3D structural manipulation.} In addressing prompts related to molecular conformation, 3DToMolo adeptly achieves the goal by manipulating 3D structures beyond functional-group-wise modifications. 
For instance, when instructed to decrease the polarity of the input molecule, 3DToMolo strategically adds a polar hydroxyl group. The added hydroxyl group spatially cancels out the dipole moment of another existing C-O bond (Figure \ref{fig:3d_polarity}(a)), resulting in a decreased total dipole moment. In another example, when tasked with increasing the polarity of a molecule with six heteroatoms, including two fluorine atoms, 3DToMolo removes highly polar C-F bonds and outputs a molecule with four heteroatoms (Figure \ref{fig:3d_polarity}(b)). This decision is based on the understanding that, in a stable conformation, the two C-F bonds contradict the dipole of pyridine ring. Thus, the replacement of C-F bonds by a hydroxyl group more aligned with the pyridine dipole in fact increases polarity. These examples underscore 3DToMolo's ability to comprehend entire molecules, including transient conformational information, a crucial aspect for precise task execution.

\begin{figure}[!htbp]
    \centering
    \includegraphics[width=0.9\linewidth]{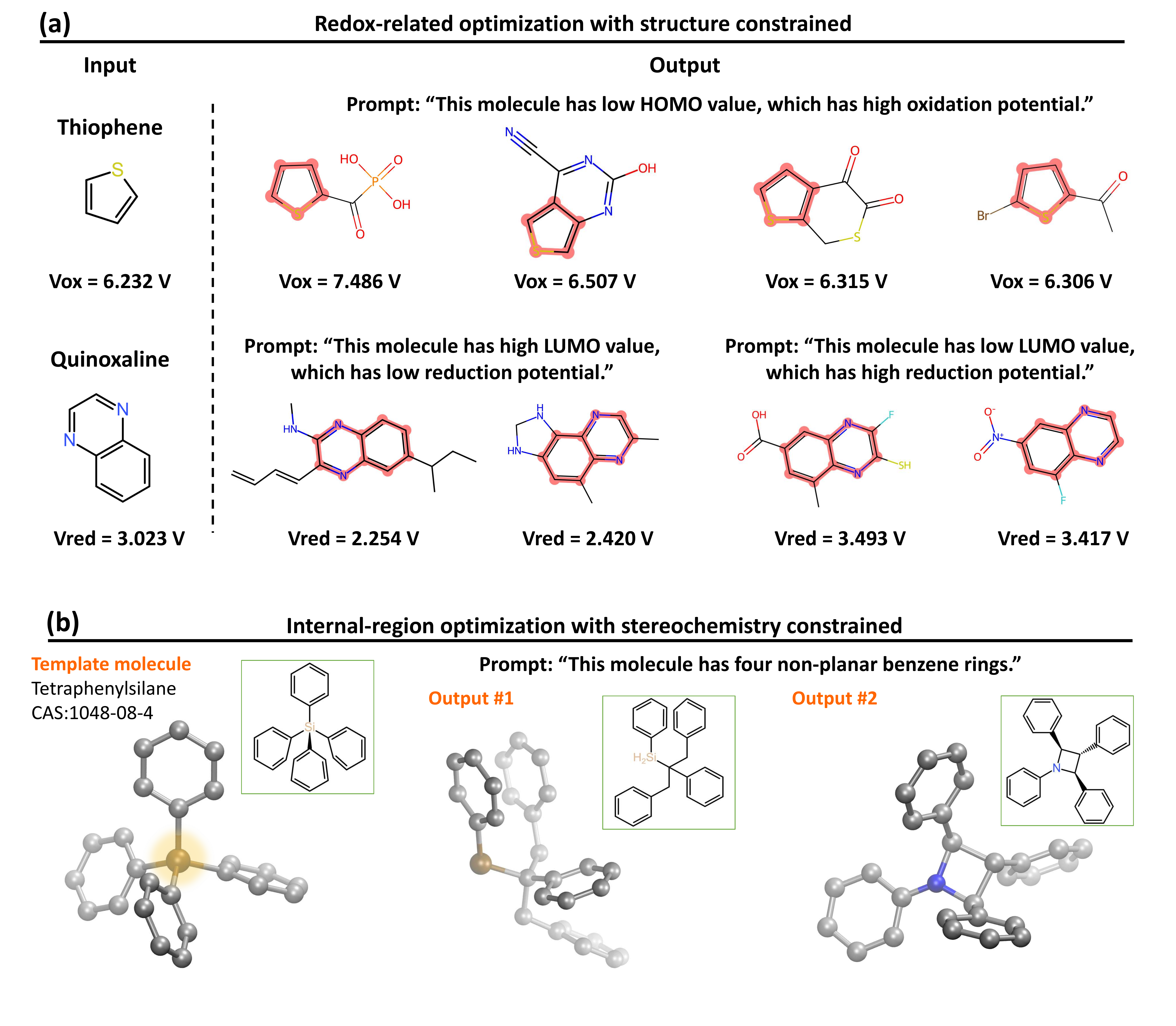}
    \caption{\label{fig:structure_constrained} Molecule optimization with structural constraints. \textbf{(a)} Redox-related prompt-driven molecule optimization with preserved skeletons. \textbf{(b)} A case study for optimization on the internal region. On the left is a reported spiro-linked $\pi-$conjugated molecule, tetraphenylsilane, used as the template. The center silicon atom (yellow-shaded area) is diffused while the four benzene rings remain fixed until final steps of the denoising process. On the right are two selected output structures whose benzene rings remain non-coplanar as desired. Hydrogen atoms are not shown for the sake of clear visualization.}
\end{figure}

\subsection{Prompt-driven Molecule optimization with Structural constraints}
While the flexible optimization scenario offers maximal optimization diversity within the chemical space, there are situations where specific substructures, as designated by experts, must be preserved. Formally, a molecule is decomposed into two disjoint parts: $M_0 \coprod S_0$, with $S_0$ representing the substructure to be protected. Consequently, we transit from recovering $p_t(M \coprod S)$ to the conditional density $p_t(M | S)$. Given that $S$ is fixed, the gradient required by the optimization process in Eq. \ref{eq: optimizing process} becomes:
$$\nabla p_t (M | S) \rightarrow \nabla_M p_t (M  \coprod S).$$
While both the variational-based prompt molecule optimization MoleculeSTM and our denoising-based approach share a common step of encoding molecules into a latent space, it is methodologically impossible to decompose the molecule into two parts in MoleculeSTM. This is because the optimization process in MoleculeSTM occurs in the latent space, which is different from our formulation.  

In the following experiments, we have carefully chosen specific physicochemical tasks to conduct a thorough examination. Specifically, we protect all core structures excluding the hydrogen atoms or other removable atoms. By only optimizing the pre-defined removable atoms, we aim to maintain the integrity of the molecular backbone, emphasizing the impact on non-hydrogen constituents, which play key roles in defining the original molecule's chemical properties and functionality.


\textbf{Redox potential related prompt-driven molecule optimization.} In the context of energy storage, enhancing the energy density of batteries necessitates elevated voltage, requiring electrolyte molecules with an expansive electrochemical window. As case studies, we aim to increase the oxidation potential and decrease the reduction potential of exemplary electrolyte molecules.

Thiophene is a common structure used in electrolyte additives for lithium-ion batteries. To augment thiophene's resistance to high voltage, we apply the prompt "This molecule has low HOMO (Highest occupied molecular orbital) value, which has high oxidation potential" while constraining all atoms except hydrogens. A comparative experiment without the prompt serves as the baseline. With the prompt, 19.5\% of the generated derivatives exhibit an increased oxidation potential, compared to 12.3\% without the prompt. Selected successful examples are presented in Figure \ref{fig:structure_constrained}(a). A similar experiment is conducted on phosphate, frequently employed in electrolytes to enhance battery stability at elevated temperatures. The success rate is 8.12\% with the prompt and 5.66\% without. To illustrate the modification of reduction potential, we use quinoxaline, pertinent to redox flow batteries. By constraining all atoms except hydrogens and employing corresponding prompts, we successfully modify the reduction potential in two directions (Figure \ref{fig:structure_constrained}(a)). In the more desirable direction of lowering the reduction potential, we achieve a success rate of 69.1\%.

Additionally, we conducted quantitative experiments a dateset of 143 common electrolyte additives \cite{redox-okamoto2018ab}. We focus on properties people generally concern about in liquid electrolytes, including the redox potential, the polarity, and the water solubility. As shown in Table \ref{tab:substructure performance}, 3DToMolo significantly outperforms the state-of-the-art baseline GPT-3.5. The major challenge for GPT-3.5 lies in the validity of output molecules, which contains problems such as incomplete rings and chemically incorrect bondings. 3DToMolo circumvents this defect, as the incorporation of 3D information makes it less likely to forget those hanging rings or bonding condition of atoms.

\begin{table*}
  \caption{The results of molecule optimization with structural constraints on the electrolyte additive dataset \cite{redox-okamoto2018ab} were evaluated using the hit ratio (\%) of property changes in each objective-oriented prompt. Statistically significant improvement (t-test over 5 different dataset splits, p-value$<$ 0.05) is highlighted with bold text. }
    \centering
  \label{tab:substructure performance}
  \begin{tabularx}{\textwidth}{m{0.5\textwidth}
    <{\centering}|X<{\centering}|X<{\centering}|X<{\centering}|X<{\centering}}
    \toprule
    \multirow{2}{*}{Prompts} & \multicolumn{2}{c|}{Validity (\%)} & \multicolumn{2}{c|}{Hit Ratio (\%)}\\
    \cline{2-5}
    & GPT-3.5 & 3DToMolo & GPT-3.5 & 3DToMolo\\
    \midrule
    This molecule has low HOMO (Highest occupied molecular orbital) value, which has high oxidation potential. & 2.867 & \textbf{41.54} & 0.2797 & \textbf{10.91} \\
    \cline{1-5}
    This molecule has high HOMO (Highest occupied molecular orbital) value, which has low oxidation potential. & 0.4196 & \textbf{42.73} & 0.06993 & \textbf{29.65} \\
    \cline{1-5}
    This molecule has low LUMO (Lowest Unoccupied Molecular Orbital) value, which has high reduction potential. & 0 & \textbf{44.55} & 0 & \textbf{25.38} \\
    \cline{1-5}
    This molecule has high LUMO (Lowest Unoccupied Molecular Orbital) value, which has low reduction potential. &  0.4895 & \textbf{42.45} & 0.4895 & \textbf{18.81} \\
    \cline{1-5}
    This molecule has high polarity. & 1.399 & \textbf{42.03} & 1.119 & \textbf{30.35} \\
    \cline{1-5}
    This molecule has low polarity. & 0.1399 & \textbf{43.15} & 00.00 & \textbf{12.17} \\
    \cline{1-5}
    This molecule is soluble in water. & 2.448 & \textbf{55.24} & 1.678 & \textbf{36.50} \\
    \cline{1-5}
    This molecule is insoluble in water. & 1.399 & \textbf{57.27} & 1.119 & \textbf{20.63} \\
    
    \bottomrule

\end{tabularx}

\begin{tablenotes}
      \small
      \item Note that the prompts we input into GPT-3.5 have subtle differences from those listed in the above table. To ensure a fair comparison, we formulated our request to GPT-3.5 as follows: "Optimize this molecule: <SMILES> to achieve <specified properties>. Optimization is restricted to hydrogen atoms only, while the rest of the input molecule remains fixed. The number of atoms (including hydrogen) added to the optimized molecule must be within the range of 1 to 10. Provide 10 different and valid results." For the compound prompts, the hit ratio is calculated based on the satisfaction of the redox potential part. In most cases, GPT-3.5 fails to meet the requirements, as the generated molecules do not conform to either the specified structural criteria or the desired properties.
\end{tablenotes}

\end{table*}


\textbf{Internal-region molecule optimization.} Modification on the internal region of a molecule is challenging \cite{du2023molecule}, as it necessitates rational linkages of fragments. This task becomes even more difficult when specific stereochemistry requirements are imposed. Here, we demonstrate the competence of 3DToMolo for such tasks via a case study on tetraphenylsilane. Non-coplanar benzene rings in tetraphenylsilane (Figure \ref{fig:structure_constrained}(b), left) is a desired structural feature for optical materials that have high refractive index and low light double-refraction (birefringence) \cite{spiroconjugate1,spiroconjugate2}. Benzene rings contribute to the strong refractive ability. The non-coplanar configuration hinders the $\pi-\pi$ stacking, preventing the formation of layering structure and thus reducing the double-refraction. To generate more candidate molecules with satisfactory configuration, we protect the benzene rings in the tetraphenylsilane molecule and diffuse the center silicon atom under the prompt "This molecule has four non-planar benzene rings". The protection is removed at final steps during the denoising process. Valid generated structures are examined by Density Functional Theory (DFT) \cite{smith2020psi4,g16} computation. Most of the structures have connected four benzene rings via the generated central motif and maintain the rings non-coplanar. Two optimized results are exemplified in Figure \ref{fig:structure_constrained}(b) and more can be found in Supplementary Figure S4. As a comparison, GPT-3.5's performance on the same task is poor either because it fails to generate required structures or because it merely conducts single-atom replacement on the silicon atom without the capability of providing more sophisticated internal structures (Supplementary Section S9). 

We focus on the prompt-driven experiments in this section. However, we note that 3DToMolo is able to handle prompt-free unconditional generation (examples in Supplementary Section S8 and Figure S5), no different from other generative models that do not incorporate LLM models.

\subsection{Hard-coded molecule optimization on appointed sites}
Precisely optimizing on pre-appointed optimization sites is notoriously difficult for latent space-based molecule representations, primarily due to the missing of exact spatial decoding. On the other hand, several machine-learning based optimization site identifiers have been proposed, specifically tailored for domain-specific tasks. Since many of these identifiers are trained on datasets where the goals are explicitly defined, and such detailed objectives may be scarce in textual representations of molecular structures. Consequently, 3DToMolo faces a hurdle in automatically identifying the desired optimization sites solely from textual prompts.
In light of this, we embark on an exploration to determine the adaptability of 3DToMolo to hard-code optimization on pre-appointed sites, establishing a comprehensive optimization pipeline. 

\begin{figure}[!htbp]
    \centering
    \includegraphics[width=\textwidth]{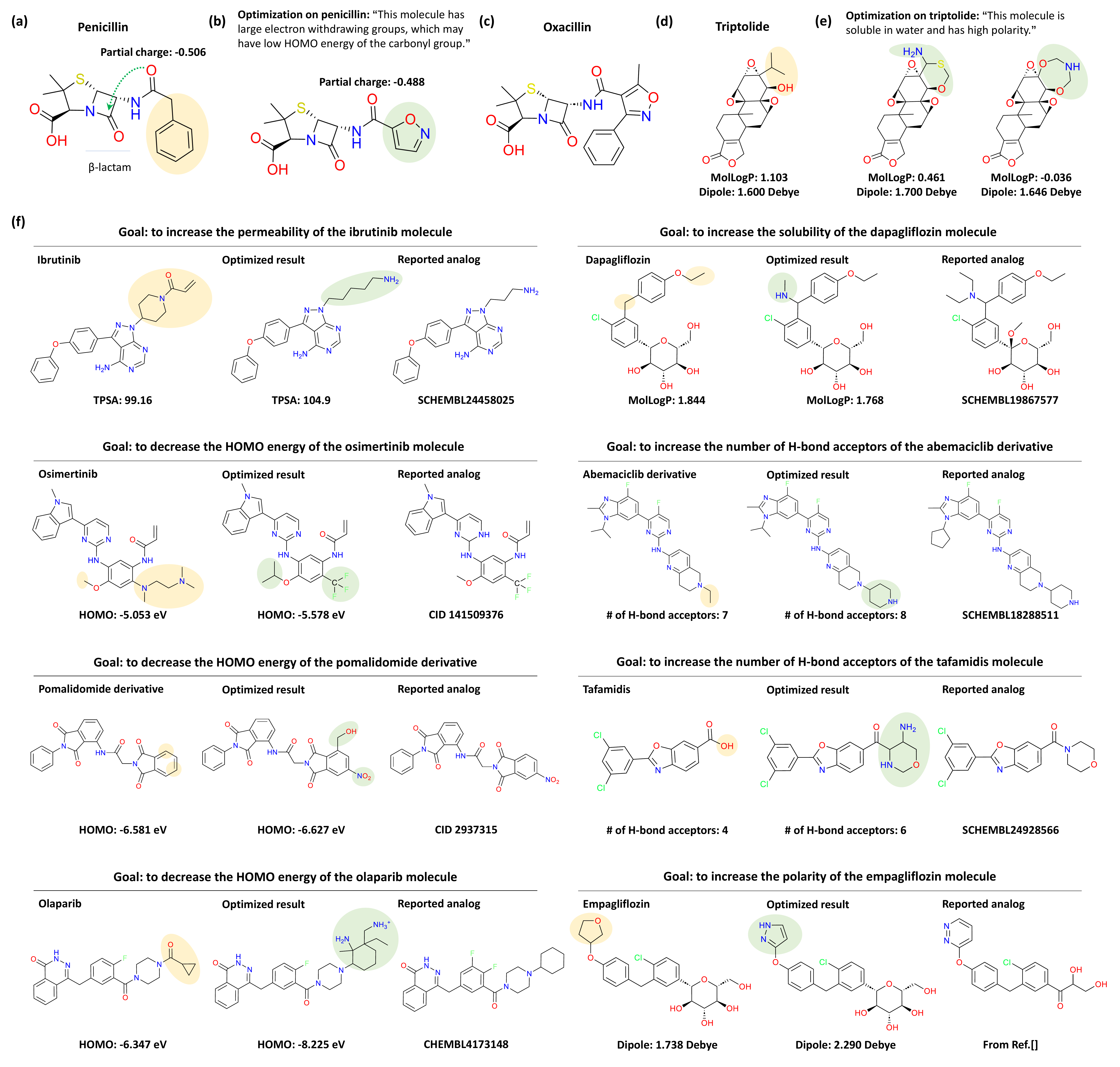}
    \caption{\label{fig:appointed_site} Drug molecule optimization on appointed sites. \textbf{(a)} The molecular structure of penicillin. The blue line marks the $\beta$-lactam ring. The dashed arrow shows the nucleophilic attack from the oxygen atom on the side chain to the $\beta$-lactam ring. The yellow region marks the functional group to be optimized. The partial charge on the nucleophilic-attacking oxygen atom is -0.506. \textbf{(b)} One representative example of our optimization results under the given prompt is shown in the figure. The green region marks the modification where the benzyl group is replaced by a more electron withdrawing isoxazole group. With the modification, the partial charge on the nucleophilic-attacking oxygen atom is successfully reduced to -0.488. \textbf{(c)} The molecular structure of oxacillin, an existing analogue to our modification. \textbf{(d)} The molecular structure of triptolide. The yellow regions mark the optimization sites.  \textbf{(e)} Two exemplary optimizations where the two modified side chains have connected. Both exhibit increases in solubility as well as in polarity, as the given prompt orders. \textbf{(f)} Selected examples in diverse drug molecule optimization tasks that have reported analogs. }
\end{figure}

From a methodological perspective, the 3D positions of the appointed sites are utilized during the hard-coded optimization process (see Supplementary Section S4 for the algorithm details). We showcase 3DToMolo's capability on two exemplary drug-related molecules, penicillin and triptolide. For penicillin, the crucial $\beta$-lactam ring \cite{penicillin1,penicillin2} is vulnerable to $\beta$-lactamase binding \cite{penicillin4} and acidic hydrolysis \cite{penicillin3} (Figure \ref{fig:appointed_site}(a)). One proposed mechanism suggests the initial nucleophilic attack by the oxygen atom from another amide group on the carbonyl group in the $\beta$-lactam ring \cite{penicillin4}. Thus, an effective strategy is replacing the benzyl group by a more electron-withdrawing functional group with substantial steric volume to impede lactamase binding and to weaken the nucleophilicity of the attacking oxygen atom. We optimize the penicillin molecule under the prompt "This molecule has large electron-withdrawing groups" while maintaining structural constraints on the entire molecule except for the benzyl group. Of the optimized structures, 23\% successfully exhibit a decrease in the electron density on the attacking oxygen atom revealed by DFT computation, with 43\% of these structures featuring a ring of at least five members, indicative of substantial steric effects. An exemplary result demonstrates the replacement of the benzyl group with an isoxazole group (Figures \ref{fig:appointed_site}(b)). Notably, the isoxazolyl series of semi-synthetic penicillins, such as oxacillin  and (Figure \ref{fig:appointed_site}(c)), has been recognized for superior resistance to acids and $\beta$-lactamases \cite{penicillin6}. A comparative test utilizing GPT-3.5 as a baseline reveals its inability to generate valid SMILES strings or preserve the core structure under varying instructions (Supplementary Section S10).

We further demonstrate the capability of 3DToMolo beyond simple side chain substitution in the optimization of Triptolide \cite{triptolide1,triptolide2} with the goal of enhancing its water solubility. Referencing to reported modified derivatives \cite{triptolide3}, we choose two adjacent sites for optimization, as depicted in Figure \ref{fig:appointed_site}(d). The remaining structure is constrained. We employ the prompt "This molecule is soluble in water and has high polarity" that has been proven to be more effective in the previous section. Within successful optimizations, we observe some connected structures between these two optimization sites (Figure \ref{fig:appointed_site}(e)), which is not achievable through sequential side chain substitution.

To provide a quantitative comparison, we construct a small task set of optimization on appointed sites in common drug molecules under given prompts. Akin to penicillin and triptolide above, modification sites and optimization strategies of each molecule are carefully chosen based on relevant studies (details in Supplementary Section S2 and Figure S1). The performance of 3DToMolo is summarized in Table \ref{tab:setting3 performance}, whereas GPT-3.5 is used as a baseline. Compared to the failure of GPT-3.5 shown in the table, 3DToMolo is more powerful in adapting the domain knowledge it has learned in diverse drug discovery tasks. Several optimized results by 3DToMolo not only meet the structural and property requirements, but also are chemically reasonable, as they resemble known molecules in existing database (Figure \ref{fig:appointed_site}(f)). These findings underscore the remarkable proficiency of 3DToMolo in selectively appointing and modifying substructures based on natural language guidance, particularly in scenarios involving complex isomeric structures and three-dimensional considerations.

\begin{table*}
  \caption{The results of molecule optimization based on specified editing positions on the small molecule drug dataset, with the evaluation measured by the hit ratio (\%) of the property changes in each objective-oriented prompt. Statistically significant improvement (t-test over 5 different dataset splits, p-value$<$ 0.05) is highlighted with bold text. }
    \centering
  \label{tab:setting3 performance}
  \begin{tabularx}{\textwidth}{
    m{0.15\textwidth}<{\centering}|m{0.2\textwidth}<{\centering}|m{0.15\textwidth}<{\centering}|X<{\centering}|X<{\centering}|X<{\centering}|X<{\centering}}
    \toprule
    \multirow{2}{*}{Drug Molecule} & \multirow{2}{*}{Objectives} & \multirow{2}{*}{Prompts} & \multicolumn{2}{c|}{Validity (\%)} & \multicolumn{2}{c}{Hit Ratio (\%)}\\
    \cline{4-7}
     & & & GPT-3.5 & 3DToMolo & GPT-3.5 & 3DToMolo\\
    \midrule
    Apixaban & Electron-donating substitution & High HOMO energy & 55 & \textbf{73} & 1 & \textbf{18} \\
    \cline{1-7}
    Empagliflozin & More hydrophilic & Large polarity & 0 & \textbf{15} & 0 & \textbf{8} \\
    \cline{1-7}
    Ibrutinib & Higher permeability & High permeability & 19 & \textbf{33} & 0 & \textbf{2} \\
    \cline{1-7}
    Dapagliflozin & Large aqueous solubility & More soluble & 0 & \textbf{35} & 0 & \textbf{13} \\
    \cline{1-7}
    Osimertinib & Electron-withdrawing substitution & Low HOMO energy & 30 & \textbf{52} &  0 & \textbf{39} \\
    \cline{1-7}
    Olaparib & Electron-withdrawing substitution & Low HOMO energy & \textbf{34} & 30 & 1 & \textbf{2} \\
    \cline{1-7}
    \multirow{2}{*}{Abemaciclib} & \multirow{2}{*}{\shortstack{More hydrogen bonding \\ interactions}} & More hydrogen bond acceptors & 20 & \textbf{42} & 0 & \textbf{21} \\
    \cline{3-7}
    & & More hydrogen bond donors & 0 & \textbf{45} & 0 & \textbf{25} \\
    \cline{1-7}
    Pomalidomide & Electron-withdrawing substitution & Low HOMO energy & 24 & \textbf{64} & \textbf{11} & 10 \\
    \cline{1-7}
    \multirow{2}{*}{Tafamidis} & \multirow{2}{*}{\shortstack{More hydrogen bonding \\ interactions}} & More hydrogen bond acceptors & 46 & \textbf{61} & 0 & \textbf{13} \\
    \cline{3-7}
    & & More hydrogen bond donors & 10 & \textbf{48} & 0 & \textbf{5} \\
    \cline{1-7}
    \multirow{2}{*}{Lisdexamfetamine} & \multirow{2}{*}{\shortstack{To control the hydrolysis \\ rate}} & High HOMO energy & 0 & \textbf{11} & 0 & \textbf{8} \\
    \cline{3-7}
    & & Low HOMO energy & 0 & \textbf{19} & 0 & \textbf{5} \\
    
    \bottomrule

\end{tabularx}
\begin{tablenotes}
      \small
      \item Note that the prompts shown in the table have been abbreviated. The prompts actually used in 3DToMolo can be found in Table \ref{tab:overall performance}, while we formulated our request to GPT-3.5 as follows: "Optimize the following molecule: <SMILES> to achieve the desired properties: <specified properties>, while ensuring that the specified substructure <substructure> remains unchanged. Provide 10 different and valid results." 
      Note in this experimental setting, we employed a multi-run approach to facilitate the generation of a total of 100 optimized results for each drug molecule. Specifically, each iteration obtained 10 distinct outputs from GPT-3.5, with a total of 10 iterations conducted. In most cases, GPT-3.5 fails to meet the requirements, as the generated molecules do not conform to either the specified structural criteria or the desired properties.
\end{tablenotes}

\end{table*}
\section{Discussion}
From a broad perspective, our text-structural optimization strategy falls within the category of multi-modality controlled molecule structural modification approaches. Given the inherent significance of 3D structures in shaping molecular properties, we employed SE(3)-equivariant graph transformers for the intricate task of encoding and decoding molecule representations. In summary, we integrated three types of modalities of a molecule: molecule graph, 3D conformers, and text descriptions. Combined with the noising-denoising 2D+3D diffusion models, 3DToMolo proves instrumental in achieving highly promising optimization outcomes. 
It allows us to optimize molecular structures not only at internal regions, enhancing flexibility, but also in pre-assigned periphery through hard-coded implementations. Notably, the achievement of such comprehensive optimization results would be unattainable without the incorporation of fine-grained 3D position imputing manipulations. 

Considering the crucial aspect of data efficiency, our design involves the decoupling of the molecule structure generation model and the text-structural alignment guidance. This separation enables us to leverage vast amounts of unlabeled structure data during the training of the structure generator. 3DToMolo proves particularly advantageous when labeled data for guiding text-structural alignment is limited.  By tapping into the abundance of unlabeled data, 3DToMolo gains a robust understanding of diverse molecular structures. Remarkably, we are able to selectively fine-tune specific aspects of the model in a low-rank manner \cite{hu2022lora}, particularly when dealing with intricate molecular geometry configurations such as binding scenarios or conformers exhibiting high energy states. 
An additional rationale for favoring diffusion models in molecule optimization lies in their global optimization approach. This stands in contrast to the local reinforced, and often greedy approach of optimizing one disconnection site at a time. The global methodology employed by diffusion models contributes to the generation of more diverse and comprehensive optimization results.
Finally, it is noteworthy that 3DToMolo distinguishes itself by not necessitating optimization trajectories as part of our training data, in contrast to traditional MCTS methods. This is attributed to the fixed nature of the noising process equation for every molecule, and only the denoising process is learned. Conversely, the manipulation of the noising process to reinforce intermediate states for synthesizability (as an example) is a plausible avenue. This adjustment holds the potential to be beneficial for generating retrosynthetic pathways for our optimized molecules.

As the field of multi-modality large neural networks advances swiftly, our research is only a preliminary attempt on harnessing the potential of multimodality information to guide the optimization of molecular structures. As an illustrative example, we utilized paired text-molecule data to train the alignment between the molecular representation $X$ and the corresponding text representation $Y$. What remains unexplored is the potential of adversarial matching between $X$ and $Y$, wherein the mapping function $G$ learns to map the distribution of $X$ to that of $Y$. This approach has the capacity to leverage unpaired text and molecule data, offering a promising avenue for further investigation. On the text side, 3DToMolo involves utilizing a pretrained Large Language Model (LLM) for extracting text embeddings. This LLM model is specifically trained by predicting the next word token based on context. While effective, a more intricate strategy involves directly training a text-molecule equivariant large model in a similar way as Emu2 \cite{sun2023generative}. This advanced approach allows for the generation of multi-modality outputs. Unlike our current approach, where the model is trained to predict molecules given the text, this more involved method also operates in reverse. It trains the model not only to predict molecules based on textual descriptions but also to predict text from molecular information. This bidirectional training scheme contributes to a more versatile and expressive representation. Moreover, extending beyond text, the inclusion of illustrations from academic papers adds another layer of informative guidance for optimizing molecules. Such image representations can also serve as valuable cues for refining and enhancing the structural optimization process.

Lastly, the uncertainty in synthesizability of generated molecules remains an unsolved problem for many deep learning models \cite{doi:10.1021/acs.jcim.0c00174}. To ensure the optimization is not only towards better properties but also better synthesizability, the model should be trained with a considerable amount of data and human expert knowledge in the synthesis domain. This is challenging because the data and the knowledge are constantly updating as new synthesis methods are discovered. Notably, although 3DToMolo does not incorporate synthesis-related data, its optimization results maintain synthetic accessibility scores \cite{SAscore} (SAscore) comparable with their inputs (Figure S6 in the Supplementary).


\section{Methods}

\subsection{Datasets}
We use PCQM4Mv2 dataset \footnote{https://ogb.stanford.edu/docs/lsc/pcqm4mv2/} to pretrain an unconditional diffusion model for modeling complex data distribution to generate new structures within the chosen chemical space. PCQM4Mv2 is a quantum chemistry dataset including  3,746,619 molecules originating from the PubChemQC project \cite{PubChemQC}. 
MoleculeSTM dataset \cite{liu2023moleculestm} with over 280K chemical structure-text pairs is used to train a text-molecule model. To better align chemical space with semantic space, we effectively incorporate 3D structure information of molecules to enhance the alignment with textual description. However, MoleculeSTM lacks 3D coordinates of molecules, thus we extract 3D information and energy-related values from PubchemQC according to the PIDs in MoleculeSTM.
Regarding downstream tasks, a novel molecule could be generated from Gaussian noise, or a molecule selected from Zinc dataset \cite{Zinc} could be optimized through applying noise and recursively denoising.


\subsection{Training details}
All methods are implemented in Python 3.9.13. PyTorch Lightning is utilized to implement a framework that maximizes flexibility without sacrificing performance at scale. All experiments were conducted on Ubuntu 20.04.6 LTS with AMD EPYC 7742 64-Core Processor, 512GB of memory, and 80GB NVIDIA Tesla V100.





\subsection{Structural optimization through diffusion}
Drawing inspiration from the demonstrated effectiveness of diffusion models in generating data from noisy inputs, such as \cite{vahdat2021score}, and image-editing applications \cite{meng2021sdedit}, we propose a 2D-3D joint
diffusion model. We aim to introduce fine-grained prompt control through the gradients derived from the contrastive loss, aligning the text-based prompt with our 2D-3D joint representation. To achieve this, our method employs a two-stage strategy. The first stage utilizes a relatively large database of molecules with diverse physicochemical properties. In the second stage, we leverage a text-molecule structure pair database for cross-modality alignment and text-guided molecule structural optimization.

\subsubsection{Denoising diffusion process}
In the first stage, we conduct pretraining of a generative 2D-3D molecular diffusion model. Following the structure of typical diffusion models, 3DToMolo encompasses two key processes: the forward process and the reverse process. To adapt these processes to 2D-3D molecular graphs, we represent the molecule $M$ as a combination of node features (atom types) $H$, an adjacency matrix $E$ (representing chemical bond edges), and 3D positions $P$, denoted as $(H,E,P)$. Specifically, suppose the molecule is composed of $n$ atoms, then $H = [H^1, \dots, H^n]$ represents the one-hot embedding in the periodic table for the $n$ atoms.  In the forward process, the original structure of the molecule undergoes a joint Markovian transition step by step. We denote the intermediate structure as $M_t: = (H_t,E_t,P_t)$, where the initial structure is denoted as $M_0$. For the 3D point cloud $P_t$, a Markov chain is implemented by incrementally introducing Gaussian noise over $T$ steps. The transition from $P_{t-1}$ to $P_t$ is described as follows:
\[
q(P_t | P_{t-1}) = \mathcal{N}(P_t; \sqrt{1-\beta_t}P_{t-1}, \beta_t I),
\]
where $t \in \{1, \dots, T\}$ denotes the diffusion step. Here, $\mathcal{N}$ represents the Gaussian distribution, and the hyperparameter $\beta_t \in (0,1)$ controls the scale of the Gaussian noise added at each step.
By leveraging the additivity property of two independent Gaussian noises, we can directly express the state $P_t$ in terms of the initial $P_0$:
\begin{equation} \label{eq: prior}
P_t = \sqrt{\bar{\alpha_t}} P_0 + \sqrt{1 - \bar{\alpha_t}} \epsilon,
\end{equation}
where $\epsilon \sim \mathcal{N}(0, I)$, and $\bar{\alpha_t} = \prod_{s=1}^t (1- \beta_s)$.
On the other hand, we treat $z_t = (E_t, H_t)$ as discrete random variables, and thus, we subject them to discrete Markov chains $Q_t$:
\[
q(z_t | z_{t-1}) = \mathcal{C}(z_{t-1} Q_t),
\]
where the transition matrix $Q_t$ represents the probability of jumping between states at diffusion step $t$, and $\mathcal{C}$ denotes the corresponding categorical distribution. Specifically, $\{Q_t\}_{0}^T$ comprises two components: $\{Q_t^H, Q_t^E\}_{0}^T$, with $Q_t^H = \alpha^z_t \cdot \mathbf{I} + \beta^z_t \cdot \mathbf{I_a} m_h$, where $m_h \in \mathbf{R^a}$ represents the marginal distribution of atom types in the training set. The $Q_t^E$ for edge diffusion is defined similarly. Finally, $\alpha^z_t$ and $\beta^z_t$ govern the noise schedule of the discrete part.
\paragraph{Stationary Distribution.}
A key ingredient for the noising process is that the stationary distribution of $q(M_t)$ is known. For example, $q(P_t)$ approaches $\mathcal{N}(0, I)$ as $t \rightarrow \infty$. Similarly, $q(H_t)$ and $q(E_t)$ follow the categorical distribution defined by $m_h$ and $m_e$ as $t \rightarrow \infty$. Since these stationary distributions are simple, we know how to sample them in a trivial way.

Next, the diffusion model learns to remove the added noises from $M_{t}$ to recover $M_{t-1}$ using neural networks. Starting from $M_T$, the reverse process gradually reconstructs the relations within the 2D and 3D representation of the molecules through the denoising transition step. Note that although the most straightforward parameterization is to directly predict $M_{t}$ given $M_{t-1}$, DDIM \cite{song2021denoising} has demonstrated that predicting the raw molecule $M_0$ is equivalent to predicting $M_{t-1}$ with the additional advantage of accelerating the generative process during inference.
\paragraph{Joint Denoising Process.}
We implement an equivariant graph transformer architecture $F_{\theta}$ inspired by \cite{vignac2023midi,du2023molecule} for predicting $M_0$ from $M_{t}$:
$$F_{\theta}(M_{t}) = (F_{\theta}^P(M_{t}), F_{\theta}^E(M_{t}), F_{\theta}^H(M_{t})).$$

For the 3D part, utilizing Eq. \ref{eq: prior} and the Bayes formula, the posterior probability is given by
\begin{equation} \label{posterior}
q(P_{t-1}| P_t, F_{\theta}^P(M_t)) \sim \mathcal{N}(\mu_t \cdot F_{\theta}^P(M_{t}) + \nu_t \cdot P_t, \sigma_t \cdot I),
\end{equation}
where $\mu_t$, $\nu_t$, and $\sigma_t$ are parameters that don't depend on the neural network.

For the 2D part, similarly, we have the discrete denoising Markov chain with the transition probability given by:
\begin{equation} \label{posterior-2d}
q(H_{t-1}| H_t, F_{\theta}^H(M_t)) \sim H_t(\Bar{Q}_t)^T \odot F_{\theta}^H(M_t)\Bar{Q}_{t-1},
\end{equation}
where $\Bar{Q}_t := Q_1 \cdots Q_t$. Following the transition states of the joint denoising process step by step, the reverse process gradually reconstructs the relations within a molecule graph and its corresponding 3D structures.

\paragraph{Optimization of the Diffusion Process.}
As demonstrated earlier, the optimization objective for learning $F_{\theta}$ is to reconstruct $M_0$ from a noised $M_t$. In practice, we found it beneficial to include a regularization term dependent on the sampled diffusion step $t$:
\begin{equation} \label{eq: loss}
\mathbf{E}_t \mathbf{E}_{q(M_t | M_{t-1})} [\lambda_t |M_0 - F_{\theta}(M_t) |^2],
\end{equation}
where $\lambda_t$ is a set of parameters depending on the noise schedule $\alpha_t$. 
From the posterior distribution (Eq. \ref{posterior} and Eq. \ref{posterior-2d}) perspective, the reconstruction loss is equivalent to optimizing an Evidence Lower Bound of the likelihood \cite{du2023flexible} of the original molecular distribution $p(M_0)$.

\subsubsection{Text-Structure Alignment}
\paragraph{Prompt guidance.}
In the previous section, we demonstrated how to reconstruct a molecule from a denoising process, laying the groundwork for meaningful molecule optimization. However, the challenge lies in guiding the denoising process to ensure that the final optimization result aligns with a given prompt. Formally, the prompt guidance denoted by $y$ is expected to influence the transition probability of the denoising process:
$$q_{\theta}(M_t | M_{t-1}) \rightarrow q_{\theta}(M_t | M_{t-1},y) = p(y,t) \cdot q_{\theta}(M_t | M_{t-1}).$$
To address this, we introduce the Clip mapping \cite{radford2021learning}, previously used in text-image alignment, to establish a connection between the text prompt and our molecular structure. The contrastive-based CLIP loss minimizes the cosine distance in the latent space between the molecule representation $X$ and a given prompt text $y$:
$$f(x,y) = \text{Clip}(x,y),$$
where $\text{Clip}$ returns the cosine distance between their encoded vectors. We utilize a pretrained molecular embedding model from \cite{du2023molecule} that maps $M$ to its vector embedding $X$. On the text side, we extract a latent embedding from a light version of pretrained large language model LLAMA-7B \cite{touvron2023llama}. During the optimization of $\text{Clip}$, the parameters of the two encoders are efficiently fine-tuned in a stop-gradient way.



Then, the amplitude of $f$ directly measures the alignment between a given molecule and its prompt text. In other words, for an original molecule embedding $X_0$, we aim for the optimized molecule $X_{\text{optimized}}$ to satisfy the condition:
$$df \cdot (X_{\text{optimized}} - X_0) > 0.$$
Thanks to the auto-differentiation technique developed by the deep learning community, obtaining the gradient $df$ with respect to the parameters of the molecular embedding model, is straightforward.

Now, we design $q_{\theta}(M_{t-1}| M_t, y)$ based on the differential $df$. Assuming $\text{Clip}$ is robustly trained, let $p(y| M_t) = \mathcal{N}(f(M_t), \sigma_y \cdot I)$, where $\sigma_t$ is a hyperparameter. Then, using the Taylor expansion:

\begin{align} \label{eq: guidance sampling}
        q_{\theta}(M_{t-1}| M_t, y) &= q_{\theta}(M_{t-1}| M_t) \cdot p(y| M_{t-1})\\
        & \approx q_{\theta}(M_{t-1}| M_t) \cdot e^{<\nabla \log p(y | X(M_t)), M_{t-1} - M_t>},
\end{align}
where $\nabla \log p(y | X(M_t))  \propto - \nabla ||y - f(M_t)||^2$. Combining the above, we set $q_{\theta}(M_{t-1}| M_t, y)$ to be:
\begin{equation}
q_{\theta}(M_{t-1}| M_t, y) \propto q_{\theta}(M_{t-1}| M_t) \cdot e^{-\lambda <\nabla_{M_t} ||y - f(M_t)||^2  , M_{t-1}>},    
\end{equation}
and the parameter $\lambda$ is introduced to control the strength of the prompt guidance.

It is worth emphasizing that all the aforementioned optimization experiments were conducted with a focus on adapting the de-noising steps. Our findings from extra experiments (detailed in Supplementary Section S6) and inherent nature of diffusion model indicate that the choice of de-noising steps is correlated with the resemblance between the optimized molecule and the initial input. In cases where minimal alterations to the input were anticipated, opting for smaller de-noising steps proved to be effective.

\paragraph{Multi-identity alignment.} To alleviate the mode-collapse issues of the global CLIP loss, we propose to utilize the method in \cite{rubungo2023llmprop} to enhance the original text embedding $y_0$ with its identity-wise embedding $y_1, \dots, y_N$ automatically extracted from the \textbf{grammar-parse tree} of the text. Note that we empirically find this technique to be beneficial for multi-objective prompt tasks. Let's take "\textit{This molecule is soluble in water, which has lower HOMO value.}". Then, the extracted identity-wise embedding is $y_1 = "\textit{soluble in water}"$ and $y_2 = "\textit{lower HOMO value}"$. The concatenated embedding $y = (y_0, y_1, y_2)$ is fed into the sampling formula Eq. \ref{eq: guidance sampling} during inference. 

\paragraph{Manifold constraint.} Drawing inspiration from the geometric explanation of the diffusion process proposed in \cite{du2023flexible}, the 3D score function $\nabla_{\theta}p_{\theta}(P_t)$ points towards the normal direction of the data manifold defined by the probability density $q(P_0)$. In the molecular scenario, this data manifold corresponds to valid molecules, constituting a low-dimensional sub-manifold within the space of all chemical graphs. For instance, the valence rule of atoms imposes a strict constraint on the topology of the graph.

However, the gradient $df$ (as seen in guidance-sampling: Eq. \ref{eq: guidance sampling}) may have negative components along the direction of $\nabla p_{\theta}(P_t)$, potentially leading to a deviation from the data manifold defined by $q(P_0)$. To address this concern, we propose subtracting the negative component from $df$ to enhance the validity of the final denoising result:
$$df \rightarrow  df - s(df, \nabla p_{\theta}(P_t))\cdot \frac{\nabla p_{\theta}(P_t)}{||\nabla p_{\theta}(P_t)||},$$
where $s(df, \nabla p_{\theta}(P_t)): = df\cdot \frac{\nabla p_{\theta}(P_t)}{||\nabla p_{\theta}(P_t)||}$.
Empirical findings suggest that incorporating the manifold constraint during sampling improves the validity of optimized results for both single-objective and multi-objective optimization tasks.

\subsection{Evaluation of physiochemical properties}
The energy minimization of optimized structures and the computation of charge distribution as well as other physiochemical properties are done by Gaussian16 \cite{g16} using B3LYP functional and 6-31G* basis set. The evaluation metric for optimized results under text prompts is the satisfactory hit ratio, which gauges whether the output molecule can fulfill the conditions specified in the text prompt (detailed in Supplementary Section S2). The high-throughput computation of redox potentials is done by a structural-descriptor-based machine learning regressor (Supplementary Section S5).

\section*{Author Contributions}

K. Z. and W. D. conceptualized the research. K. Z. wrote the code. K. Z., W. D., and Y. L. designed the downstream tasks. K. Z., Y. L., and G. W. analyzed the data and results. K. Z., W. D., and Y. L. drafted the manuscript. B. W., Y. R., and X.-Y. Z. supervised the experimental process.

\section*{Competing Interests}

The authors declare no competing interest.

\section*{Data Availability}

All datasets used in this document are publicly available. The PCQM4MV2 dataset including optimized structures, various properties and 3D information of all the 3,378,606 training molecules is available at acs.jcim.7b00083: \url{https://doi.org/10.1021/acs.jcim.7b00083}.
The MoleculeNet dataset is available at C7SC02664A: \url{https://doi.org/10.1039/C7SC02664A}.
The MoleculeSTM dataset is available at s42256-023-00759-6:  \url{https://doi.org/10.1038/s42256-023-00759-6}.
The PubChemQC dataset is available from \url{https://nakatamaho.riken.jp/pubchemqc.riken.jp/}. We used the HOMO, LUMO, and HOMO-LUMO gap from PubChemQC dataset, which was obtained at B3LYP/6-31G* level. 
The redox dataset is available in the Supplementary of acsomega.8b00576: \url{https://doi.org/10.1021/acsomega.8b00576}.
\section{Acknowledgments}

This work was supported by the National Natural Science Foundation of China (NSFC) (Grant 62376265). And we thank Chunngai Hui for the insightful discussions in organic chemistry and small-molecule drug related tasks.









\appendix


\end{document}

\renewcommand{\thefigure}{S\arabic{figure}}
\renewcommand{\thetable}{S\arabic{table}}
\renewcommand{\theequation}{S\arabic{equation}}
\renewcommand{\thesection}{S\arabic{section}.}

\setcounter{figure}{0}
\setcounter{table}{0}
\setcounter{equation}{0}
\setcounter{section}{0}

\maketitle

\section{More Related Works}

In recent years, the scientific discourse surrounding molecule discovery has burgeoned, driven by a collective effort to confront the intricate challenges inherent in identifying novel compounds endowed with specific and desired properties. Within this expansive domain, one prominent line of research focuses on generative models, such as variational autoencoders (VAEs) \cite{gomez2018automatic, kusner2017grammar, nakata2018molecular, simonovsky2018graphvae} and generative adversarial networks (GANs) \cite{goodfellow2020generative, prykhodko2019novo, gomez2018chemgan, de2018molgan}, which leverage deep learning \cite{krishnan2021novo, arus2019randomized, bagal2021molgpt, mahmood2021masked, gupta2018generative, li2021structure} techniques to generate new molecules. These models have demonstrated promising results in generating diverse and chemically valid molecules. By formulating the molecule optimization problem as a sequence-to-sequence or graph-to-graph translation problem,  \cite{He2021,Hoffman2022} also utilizes molecular autoencoders as the backbone model for purely 2D molecule optimization. Another approach entails the employment of Reinforcement Learning (RL) algorithms to iteratively optimize molecular structures guided by predefined objectives. RL-based methods \cite{atance, popova2018deep, olivecrona2017molecular,putin2018reinforcement,you2018graph, segler2018planning} have shown potential in optimizing drug-like properties and exploring chemical space efficiently. However, the diversity of molecules is still a significant challenge in the realm of molecular generation, and both VAE and GAN based models encounter limitations in effectively addressing this issue. The inherent nature of these models often results in generated molecules lacking the desired diversity and complexity due to the vast chemical space and intricate nature of molecular structures. However, a promising solution emerges through the adoption of diffusion model-based molecular generation. Unlike traditional approaches, diffusion models excel at capturing the stochastic nature of molecular transformations, allowing for the generation of diverse molecular structures while maintaining chemical validity. By embracing the principles of diffusion processes, these models offer a more nuanced and accurate representation of molecular diversity, thereby presenting a potential breakthrough in advancing the capabilities of molecular generation methodologies. Through manipulating the diffusion process, it becomes possible to expand more diverse molecules with desired properties, which offers valuable insights and inspiration for scientific research.

Along with the extensive data accessible, researchers have turned their attention to leveraging large-scale, unlabeled molecular datasets to pretrain deep neural networks. The paradigm of unsupervised pretraining on a vast amount of molecular data has emerged as a cornerstone in the quest for enhanced representation learning.
Nevertheless, this paradigm encounters prominent challenges. 1) Unsupervised pretraining, while advantageous in acquiring knowledge of chemical structures devoid of supervised annotations, achieves this through the reconstruction of masked topological \cite{hu2019strategies} or geometric substructures \cite{Liu2022MolecularGP}. In contrast to the familiar territory of supervised learning with labeled data, a significant challenge arises in generalizing to unseen categories and tasks without the luxury of such labeled instances or the refinement afforded by fine-tuning. 2) The success of multimodal pretraining in various domains such as image captioning \cite{karpathy2015deep, lu2019vilbert}, video analysis \cite{wang2016temporal, miech2019howto100m}, and natural language processing \cite{liu2023multimodal} has shown that the synergy between different modalities enables the model to capture intricate relationships and patterns that may be overlooked in unimodal settings. By providing models with a rich tapestry of information, multimodal pretraining not only enhances their ability to comprehend and generate content across modalities but also fosters a more nuanced understanding of the underlying semantic connections between different types of data. 
Previous studies primarily concentrated on string representations of molecules \cite{kusner2017grammar, gomez2018automatic, sanchez2018inverse, segler2018generating}, such as SMILES (Simplified Molecular Input Line Entry System) \cite{weininger1988smiles}, without effectively incorporating the structural information inherent in molecules. And there are also many works centering on two-dimensional molecular graphs \cite{duvenaud2015convolutional, liu2019n, liu2023moleculestm, zeng2022deep}. Unsupervised approaches that incorporate both two-dimensional and three-dimensional geometric structures in both pretraining and downstream tasks remain less explored. It is noteworthy that a concurrent study \cite{Li2024Towards3M} also explores the integration of 3D and textual data for molecular properties and caption learning. However, a significant distinction lies in our model's development of a robust diffusion model as a versatile decoder to enhance the optimization of molecular structures. From a broader perspective, we assess our text-3D approach in a zero-shot optimization scenario and propose future investigation into extending zero-shot to few-shot optimization, as suggested by \cite{liu2024conversational}, to further enhance performance. 


\section{Background for baseline models and optimization tasks}
\paragraph{Representative state-of-art machine learning baselines.} The baselines we used in this work include:
\begin{itemize}
    \item \textbf{MoleculeSTM:} A multi-modal model, which enhances molecule representation learning through the integration of textual descriptions. On zero-shot text-based molecule optimization tasks, the effectiveness of MoleculeSTM has been confirmed compared to the existing methods. However, this approach has overlooked the 3D geometric information. Moreover, this method suffers from less diverse generation results as it lacks flexibility during the generation process.

    \item \textbf{GPT3.5:} With its immense language processing capabilities and a broad understanding of chemistry concepts, has the potential to revolutionize molecule optimization. GPT3.5 can learn the patterns and relationships inherent in molecular structures by training on vast amounts of chemical data. By interacting with the model, we could input specific molecular structures or describe desired modifications, and GPT3.5 can generate novel molecules with desired properties. However, it is important to note that GPT3.5, as powerful as it is, should be used in conjunction with human expertise and validation. The generated molecules may not follow chemical rules or lack of sophisticated modifications.
    \item \textbf{Galactica:}  A versatile scientific language model, extensively trained on a vast repository of scientific text and data. Its capabilities extend beyond standard scientific NLP tasks, excelling in sophisticated endeavors like predicting citations, conducting mathematical reasoning, predicting molecular properties, and annotating proteins at a remarkable level of proficiency.
    
\end{itemize}

\paragraph{The 18 optimization tasks in Table 2 in the main article.} The highest occupied molecular orbital (HOMO) energy and the lowest unoccupied molecular orbital (LUMO) energy are two important measures of the reaction capability and activity of a molecule. Common chemical reactions usually involve the electron transfer from the HOMO of one molecule to the LUMO of another molecule with approximate energy value. Hence, a higher HOMO and/or a lower LUMO correspond to a larger reaction tendency, and vice versa. The HOMO-LUMO gap accounts for the light absorption of a molecule, as photons with energy equal to the gap can excite electrons from the HOMO to the LUMO.

Polarity determines the intermolecular interaction strength and is quantified by the dipole moment, which is a measure of the separation of the charge centers of electrons and atom nuclei. The total polarity of a molecule, determined by the sum of polar vectors of each heteroatomic bond, can be tuned by adjusting individual bond polarity or modifying the spatial displacement of these polar bonds. A larger polarity leads to stronger interactions and hence higher freezing and boiling points of a matter. Polarity also governs the solubility, as a molecule tends to dissolve into solvents with similar polarity (the like dissolves like rule). Water solubility and permeability of a molecule are respectively determined by its interaction strength with water, which is polar, and lipid membranes, which are non-polar. Therefore, both can be tuned via modifications of the polarity. Specifically, water solubility is also determined by the hydrogen-bond forming capability of the molecule. A hydrogen bond is formed between a highly electronegative atom (fluorine (F), oxygen (O), and sometimes nitrogen (N)), termed \textit{hydrogen bond acceptor}, and a hydrogen atom bonded with another highly electronegative atom, termed \textit{hydrogen bond donor}. Other than water solubility, hydrogen bonds play a significant role in various biochemical interactions. 

Lastly, drug-likeness prediction assesses the bioactivity of a molecule qualitatively and is important for the virtual screening of drug candidates.

\paragraph{Evaluation of optimization performance.} The evaluation metric is the satisfactory hit ratio. Given the input molecule $M_0$ and a text prompt $y$, optimized molecule $M_y$ is outputted by 3DToMolo. Then we use the hit ratio to measure if the output molecule can align with the conditions in the text prompt

\begin{equation}
    hit(M_0, y) = \begin{cases}
    1, valid(M_y) \cap satisfy(M_0, M_y, y) \\ 
    0, otherwise 
    \end{cases} 
    , hit(y) = \frac{\sum^{N}_{i=1}hit(M_0^i, y)}{N},
\end{equation}
where $N$ is the total number of molecules to be optimized, $valid()$ is the chemical test for measuring whether a molecule is valid or not, and $satisfy()$ is the condition calculation function. It is task-specific, the logarithm of partition coefficient (LogP), quantitative estimate of drug-likeness (QED), and topological polar surface area (tPSA) are used as the proxies to evaluate the molecule solubility \cite{leo1971partition}, drug-likeness \cite{bickerton2012quantifying}, and permeability \cite{ertl2000}, respectively.  The count of hydrogen bond acceptors (HBA) and hydrogen bond donors (HBD) are calculated explicitly. The value of Homo, Lumo, Homo-Lumo gap, and polarity are obtained through DFT. For single-objective optimization, a successful hit represents the measurement difference between the input molecule and output molecule above a certain threshold (we take 0). For multiple-objective property-based optimization, a text prompt should compose multiple properties' descriptions and a successful hit needs to satisfy all the properties simultaneously.

\paragraph{Binding-affinity-based molecule optimization.} We conducted binding-affinity-based molecule optimization to validate that the ligands generated by 3DToMolo could bind with the receptor in lower docking scores. Note that different from traditional pocket-ligand joint geometry modeling methods \cite{caumes2017investigating}, 3DToMolo doesn't require training another neural network model on the pocket side, which also demonstrates the flexibility of the text-prompt driven molecule optimization. The assay description serves as the basis for binding-affinity-based optimization, with each assay representing a distinct binding affinity task. A notable instance is exemplified by ChEMBL 1613777 \cite{mendez2019chembl} ("This molecule is tested positive in an assay that are inhibitors and substrates of an enzyme protein. It uses molecular oxygen inserting one oxygen atom into a substrate, and reducing the second into a water molecule.").

\paragraph{Redox potentials.} Redox potential, synonymous with reduction-oxidation potential, quantifies a substance's propensity to lose or gain electrons in a chemical redox reaction. This property holds significance in diverse fields such as chemical synthesis, catalysis, and energy storage. In the context of energy storage, enhancing the energy density of batteries necessitates elevated voltage, requiring electrolyte molecules with an expansive electrochemical window.

\paragraph{Optimization on penicillin, triptolide, and other drug molecules.} The $\beta$-lactam ring is crucial for the antibiotic function of penicillin \cite{penicillin1,penicillin2}. However, this ring is vulnerable to hydrolysis in acidic environments \cite{penicillin3}. A proposed hydrolysis mechanism suggests the initial nucleophilic attack by the oxygen atom from another amide group on the carbonyl group in the $\beta$-lactam ring \cite{penicillin4}. To enhance penicillin's acid resistance, reducing the electron density on the attacking oxygen atom and weakening its nucleophilicity are crucial. Additionally, the $\beta$-lactam serves as the binding site for $\beta$-lactamases, enzymes that confer resistance to $\beta$-lactam drugs by hydrolyzing the amide bond in the $\beta$-lactam ring \cite{penicillin5}. Thus, an effective strategy is replacing the benzyl group by a more electron-withdrawing functional group which exhibits substantial steric hindrance to impede lactamase binding.

Triptolide has been extensively investigated for its mechanism of action and pharmacological activities. It exhibits significant benefits, including anti-cancer, anti-rheumatoid, anti-inflammatory, anti-Alzheimer’s effects, and more \cite{triptolide1}. Despite these promising attributes, neither triptolide nor its analogs have gained approval as drugs, primarily due to poor water solubility \cite{triptolide2}. Therefore, our objective is to enhance the water solubility of triptolide.

\begin{figure}[H]
    \centering
    \includegraphics[width=\linewidth]{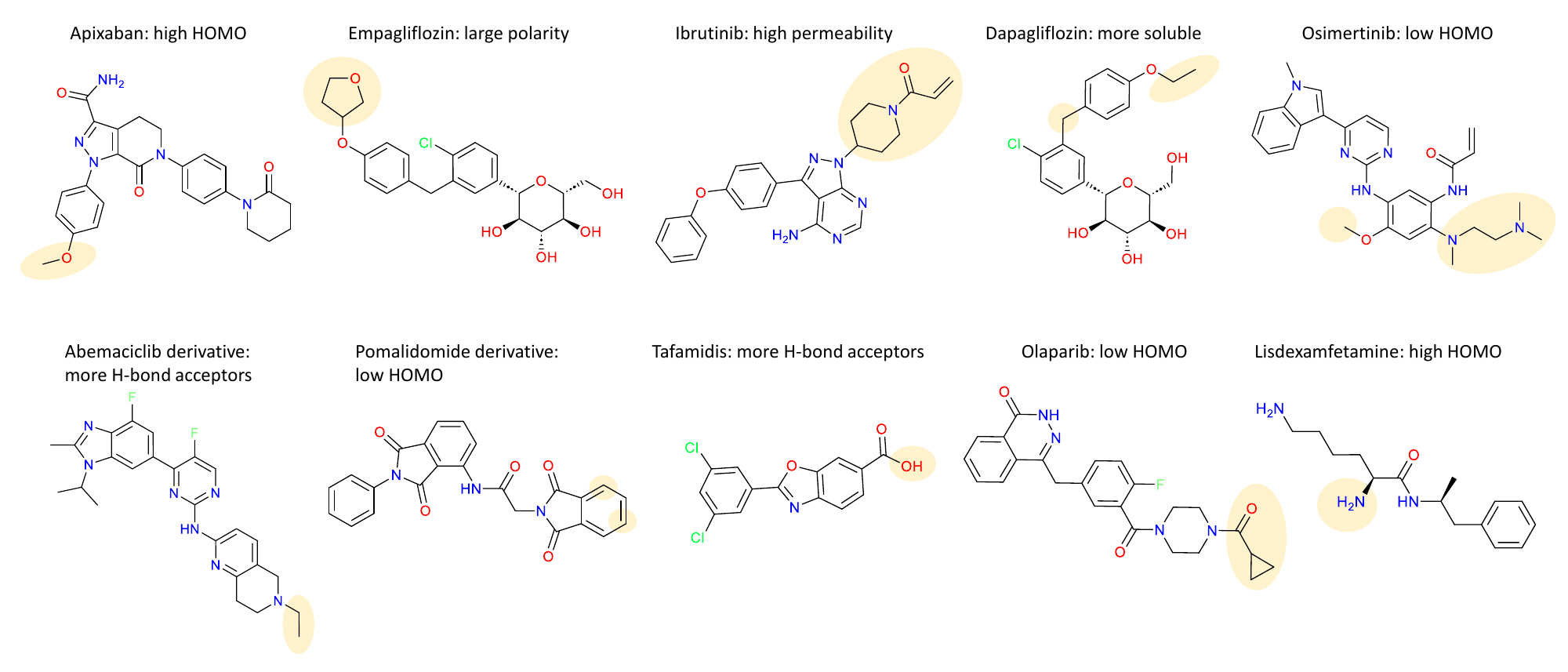}
    \caption{Chemical structures, optimization positions, and optimization prompts of selected drug molecules. The optimization is only applied on the part marked as yellow, whereas the rest of the molecule is reserved.}
    \label{fig:drugtask}
\end{figure}

More drug molecules that form a mini dataset for the hard-core optimization task with structural constraints (see Table 4 in the main article) are selected from the best selling small molecule drugs in 2023 \cite{topdrugs}. Their chemical structures, optimization positions, and optimization strategies are shown in Figure \ref{fig:drugtask}. The reasoning behind the optimization tasks are as follows:
\begin{itemize}
    \item \textbf{Apixaban} is an anticoagulant and a blood thinner that prevents blood clots and stroke. It is revealed that replacing the methoxyl group by electron-withdrawing groups significantly reduces the anticoagulant activity \cite{apixaban1,apixaban2}. Thus, we aim to optimize the methoxyl group by a more electron-donating group. This aim is aligned with and can be achieved by the \textit{high HOMO} prompt.

    \item \textbf{Empagliflozin} is a diabetes and anti-heart failure medicine. It is found that the hydrophilic substituents on the distal aromatic ring of the molecule are beneficial for maintaining or even improving cardioprotective activity and reduce the toxicity of compounds \cite{empagliflozin}. Generally, large polarity leads to good hydrophilicity, hence the prompt.
    
    \item \textbf{Ibrutinib} is a kinase inhibitor that is used to treat various blood cancers. One of the goals in studies on ibrutinib and its analogs is to improve the permeability of the molecule \cite{ibrutinib}.

    \item \textbf{Dapaglifozin} helps control blood sugar levels is a medication used to treat type 2 diabetes. As dapagliflozin is crystalline in nature, its poor oral bioavailability stems from the low aqueous solubility \cite{dapagliflozin}. As a result, we tend to improve the solubility of the molecule.

    \item \textbf{Osimertinib} is a medicine used to treat lung cancer. Studies reveal that the substitution of side chains by groups with strong electronegativity, such as fluorine, is an effective method to improve the affinity to the target site \cite{osimertinib}. This aim is aligned with and can be achieved by the \textit{low HOMO} prompt.

    \item \textbf{Olaparib} is a Poly (ADP-ribose) polymerases-1 inhibitor used to treat various kinds of cancers. It is revealed that substitutions on the carbonyl side chain by electron-withdrawing groups generally enhance the inhibition strength \cite{olaparib}. To introduce electron-withdrawing groups, we utilize the \textit{low HOMO} prompt.

    \item \textbf{Abemaciclib} is a medication that blocks the enzymes CDK4 and CDK6 and hence inhibits the growth of breast cancer cells. Research on one of the abemaciclib derivatives (see Figure \ref{fig:drugtask}) finds that the hydrogen bonding interactions between the tetrahydro-naphthyridine moiety and the residues in CDK4/6 is crucial for the blocking \cite{Abemaciclib}. Therefore, we aim to enhance the interaction by introducing more hydrogen-bond acceptors/donors into the molecule.

    \item \textbf{Pomalidomide} is an antineoplastic medication used to treat multiple myeloma and Kaposi sarcoma. Studies show that electron-withdrawing substitutions on the indole moiety in one of the pomalidomide derivatives (see Figure \ref{fig:drugtask}) is beneficial for bioactivity \cite{pomalidomide}. To introduce electron-withdrawing groups, we utilize the \textit{low HOMO} prompt.

    \item \textbf{Tafamidis} is a medication used to treat cardiomyopathy and peripheral neuropathy caused by transthyretin (TTR) amyloidosis. Tafamidis stabilizes TTR and inhibits its dissociation into monomers by binding to the TTR tetramer via hydrogen bonds, where the carboxylate group of tafamidis plays a key role \cite{tafamidis}. Therefore, optimizations on the carboxylate position that allow the formation of more hydrogen bonds can be helpful.

    \item \textbf{Lisdexamfetamnine} is used to treat neurological diseases, primarily for attention deficit hyperactivity disorder and binge eating disorder. Lisdexamfetamnine is an inactive prodrug which works only after being enzymatically hydrolyzed into active agent, dextroamphetamine, in the blood \cite{lisdexamfetamine}. A fine control of the release time is important for such a prodrug, which chemically is related to the hydrolysis rate \cite{lisdexamfetamine}. Akin to the penicillin optimization task, we aim to affect the hydrolysis rate via introducing substituents on neighboring positions that tune the HOMO energy of the molecule.
    
\end{itemize}

\section{SE(3) equivariant graph transformer}
From the reconstruction training objective Eq. (6) in the main article, the neural network $F_{\theta}$ is responsible for mapping the noised molecule $M_t$ to the original molecule $M_0$. We adopt a graph transformer architecture due to its capacity for capturing complex graph information \cite{du2023molecule,vignac2023midi,pmlr-v162-du22e}. The inputs consist of the nuclear charges $H_i$, a dense edge matrix consisting of bond types (where the 'null' bond is also denoted as an edge type), and the positions $P_i \in \mathbf{R}^3$. The invariant inputs $H$ and $E$ are fed into an embedding layer to create a high-dimensional representation of atoms and bonds, denoted by $A$ and $W$. The embedding of equivariant $P$ will be introduced in the following paragraph.

\begin{figure}[!htbp]

    \centering
    \begin{minipage}[c]{\textwidth}
    \centering
    \includegraphics[width=0.9\textwidth]{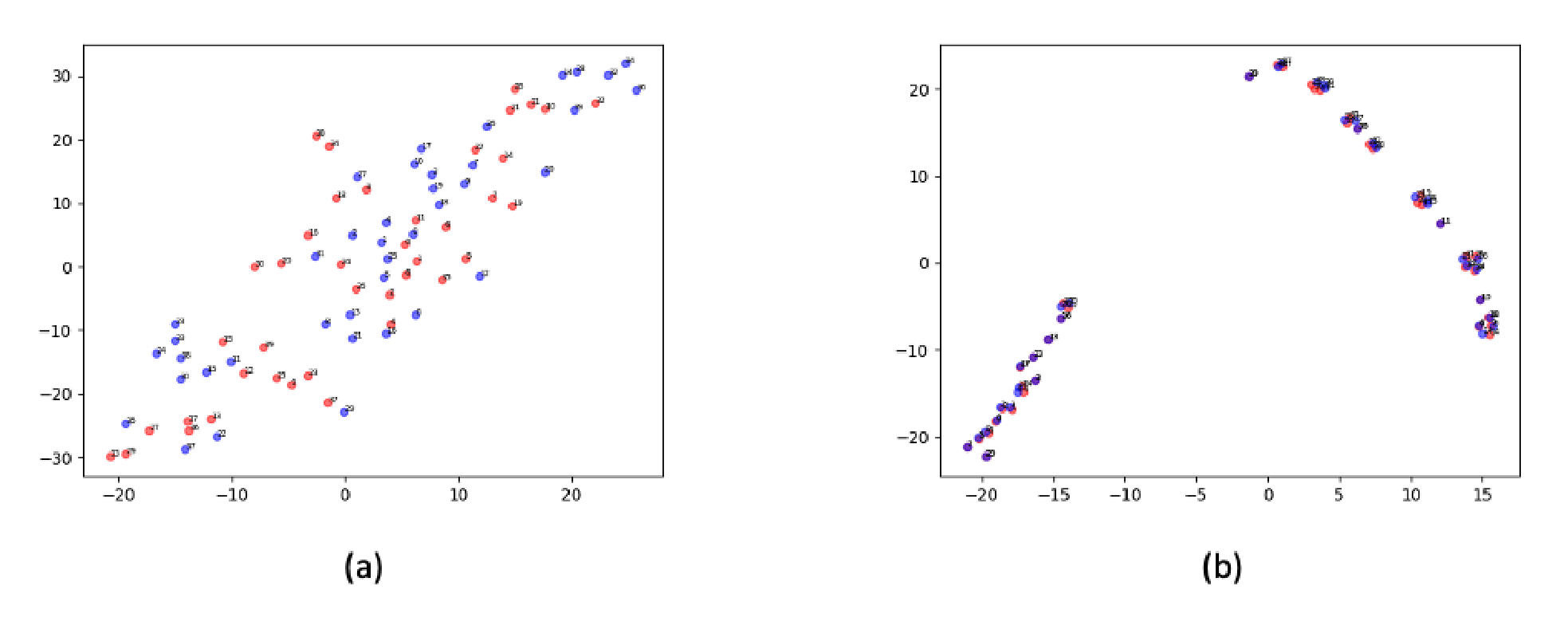}
    \end{minipage}

    \caption{Comparison between 3DToMolo and EDM \cite{hoogeboom2022equivariant} on the chirality discrimination. T-SNE projects the representations of molecules with different noise-adding steps (we take 0 to 200 at an interval of 5) into a 2D representation. In the scatter plot, the red and blue dots represent a pair of left and right-handed chiral molecules, respectively.  Figure (a) is the scatter plot for our 3DToMolo and (b) is for EDM. The dispersion or overlap of scatter points implies the capability of model for distinguishing left and right-handed chirality molecules. We could observe that the backbone model of 3DToMolo could distinguish between left and right-handed molecules with high precision, which is not achievable for EDM.
}
    \label{fig:SI_chirality} 
\end{figure}

\paragraph{SE(3) Embedding.}
A notable property of molecular 3D conformers data distributions is the SE(3) symmetry \cite{batzner20223} and the reflection anti-symmetry (Figure \ref{fig:SI_chirality}). Therefore, our embedding of equivariant $P$ should also be rotationally and translationally invariant and reflect the difference between chiral molecules. Furthermore, the 3D output $F_{\theta}^P(M)$ should transform equivalently with respect to the $SE(3)$ group. A unified framework for encoding and decoding equivariant quantities is through scalarization and tensorization \cite{pmlr-v162-du22e}, which depends on a set of equivariant frames. Two common choices of equivariant frames are vector frames and spherical harmonics. In this work, we apply edge-based vector frames that are easily merged with 2D invariant features from $A$ and $W$.

Let $\textbf{p}_i$ and $\textbf{p}_j$ be the two atom positions connected by bond $e_{ij}$. Then, the SE(3) symmetric, reflection anti-symmetric frame is defined by:
\begin{equation} 
\label{eq:edge frame}
\mathcal{F}_{ij} =(\mathcal{F}_{ij}^1,\mathcal{F}_{ij}^2,\mathcal{F}_{ij}^3) :=\left(\frac{\textbf{p}_i - \textbf{p}_j}{||\textbf{p}_i - \textbf{p}_j||}, \frac{\textbf{p}_i \times \textbf{p}_j}{||\textbf{p}_i \times \textbf{p}_j||}, \frac{\textbf{p}_i - \textbf{p}_j}{||\textbf{p}_i - \textbf{p}_j||} \times \frac{\textbf{p}_i \times \textbf{p}_j}{||\textbf{p}_i \times \textbf{p}_j||}\right).
\end{equation}

Then, fixing an equivariant frame $\mathcal{F}$, we define the \textbf{Scalarization} operation with respect to an equivariant vector $\textbf{v}$ as:
$$S(\textbf{v}, \mathcal{F}) : = (\mathcal{F}^1 \cdot \textbf{v}, \mathcal{F}^2 \cdot \textbf{v}, \mathcal{F}^3 \cdot \textbf{v}),$$
such that the output is a tuple of invariant scalars. Reversely, we define the \textbf{Tensorization} operation with respect to a tuple of scalars $a=(a_1,a_2,a_3)$ as:
$$T(a, \mathcal{F}) : = (a_1 \mathcal{F}^1, a_2 \mathcal{F}^2, a_3 \mathcal{F}^3),$$
such that the output is an equivariant vector. By iterating these two operations through different layers of the neural network, we can reform standard graph transformers into equivariant architectures.

\paragraph{Attention-Based Update Module.}

Following the standard transformer architecture \cite{NIPS2017_3f5ee243}, we implement a residual structure and layer normalization between different update blocks to prevent gradient exploration and vanishing. Within each layer, we utilize the feature-wise attention mechanism that aggregates 2D and 3D information between atoms. Unlike the standard method of calculating attention coefficients, 3DToMolo leverages the equivariant 3D features from the position matrix $P$, thereby merging 3D information and 2D information during updating.

Let $A^{l-1} \in \mathbf{R}^{N \times d}$, $W^{l-1} \in \mathbf{R}^{N \times N \times d}$ be the atom-wise features and edge-wise features at layer $l-1$, and let $P^{l-1} \in \mathbf{R}^{N \times 3}$ denote the equivariant vector features matrix at layer $l-1$ ($P^0$ represents the initial position matrix $P$). Building on the previous section, we have edge-wise equivariant frames $\mathcal{F}_{ij}$ for each edge $e_{ij} \in E$. We then scalarize the equivariant positions $p_i^{l-1}$ and $p_j^{l-1}$ to obtain two scalar tuples: $S(\textbf{p}_i^{l-1}, \mathcal{F}_{ij})$ and $S(\textbf{p}_j^{l-1}, \mathcal{F}_{ij})$. The self-attention message passing is given by:
\begin{align}
\alpha_{ij}^{l-1} & =  \text{Softmax} \{\frac{(W_{Q} a_i^{l-1}) \cdot (W_{K} a_j^{l-1})}{\sqrt{d}} + \phi^{l-1}( S(\textbf{p}_i^{l-1}, \mathcal{F}_{ij}) || S(\textbf{p}_j^{l-1}, \mathcal{F}_{ij}) || w_{ij}^{l-1}) \}, \\
\Delta x^{l-1} &= \sum_j \alpha_{ij}^{l-1} \cdot (W_V a_j^{l-1}), \\
a_i^l &= \phi_a (\Delta x^{l-1}) + a_i^{l-1},\\
w_{ij}^l &= \phi_w (\Delta x_i^{l-1} || \Delta x_j^{l-1}) + aw_{ij}^{l-1},
\end{align}
where $||$ denotes the concatenation operation between tensors. On the other hand, the equivariant $P^l$ is updated through tensorization:
$$\textbf{p}_i^l = \textbf{p}_i^{l-1} + \sum_j T(\phi_{\textbf{p}} (\alpha_{ij}^{l-1}), \mathcal{F}_{ij}).$$
For the normalization update module, we apply standard layer normalization on scalar $A$ and $W$, and apply equivariant $E_3$ normalization \cite{vignac2023midi} to the position feature $P$.

Our updating formulas mix different types of features through self-attention, and we can extend the capacity of 3DToMolo by incorporating SMILES representation through cross attention. As SMILES and molecule graphs are merely different representations of molecules, this flexibility allows us to take advantage of powerful pretrained SMILES encoders like REINVENT4 \cite{Loeffler2024}.

\section{Multi-Modality Optimization of 3DTomolo}  
\label{sup: optimizing algorithm}

\begin{figure}[!htbp]
    \centering
    \begin{minipage}[c]{\textwidth}
    \centering
    \includegraphics[width=0.6\textwidth]{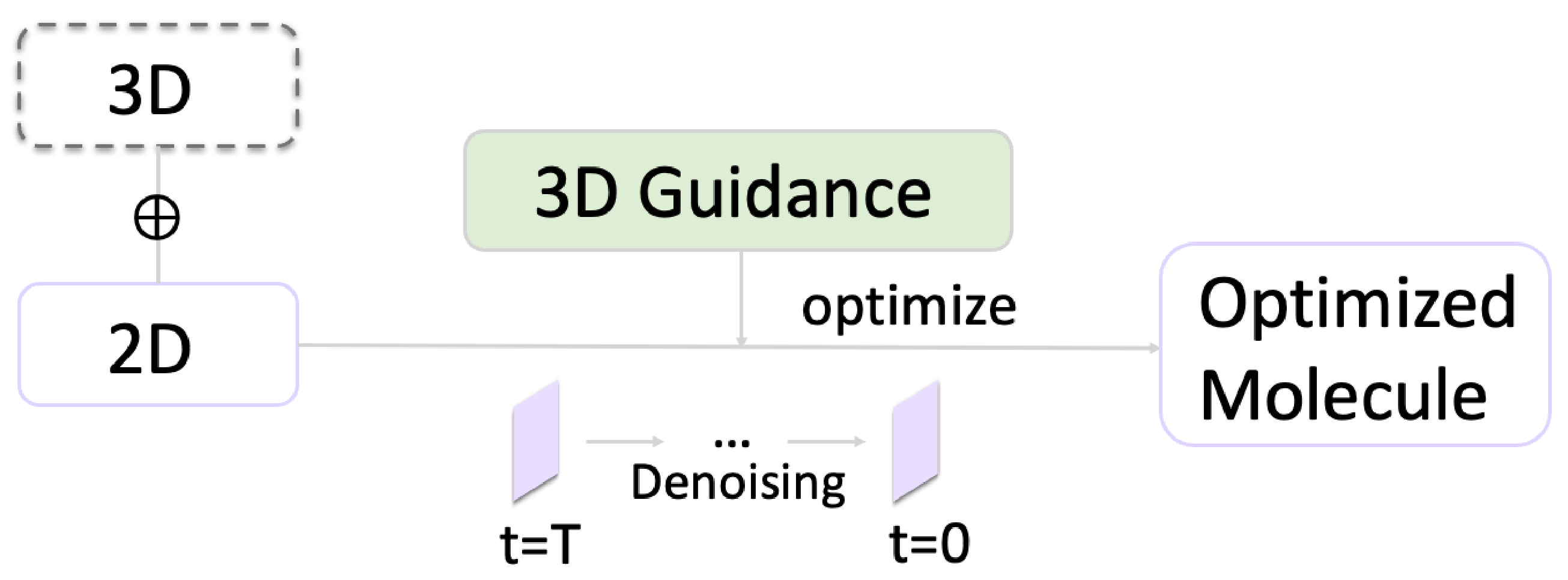}
    \end{minipage}

    \caption{Method Sketch.
}
    \label{fig:SI_arch} 
\end{figure}

In our 3DToMolo framework (Figure \ref{fig:SI_arch}), besides the target molecule probability density $p_0(\text{mol})$, we introduce a family of noised distributions $p_t(\text{mol})$ whose stationary distribution is $p_0(\text{mol})$. Therefore, $p_t(\text{mol})$ can be considered as a blurred encoding of the raw molecule for optimization. Essentially, if we initiate the denoising process from a $p_t(\text{mol})$ with a value of $t$ close to zero, the resultant molecule will resemble the raw molecule. Formally, let $X_t$ denote the noised molecule derived from $X_0$, and let $df$ represent the differentiable guidance from a specific prompt (see Eq. (7) and Eq. (8) in the main article). The optimization process can be illustrated as:
$$X_0 \xrightarrow{\text{Noising}} X_t \xrightarrow{df \text{guided Denoising}}  X_0^f.
$$
Although the noising process blurs high-resolution information, it preserves the skeleton information. Furthermore, we can parameterize the noising process \cite{du2023flexible} to control which parts of the information are preserved.

Another method for achieving fine-grained optimization of the molecule involves noising only a sub-structure $S_0$ of the molecule, ensuring that the de-noising process does not disrupt the fixed sub-structure. Formally, we decompose $X_0 = M_0 \coprod S_0$, and we only optimize the $M_0$ part:
$$X_t \xrightarrow{Noising}  M_t \coprod S_0 \xrightarrow{df \text{guided Denoising}} X_0^f: = M_0^f \coprod S_0. $$
Note that we don't assume that the position of $S_0$  is near the center of the molecule. Thus we need to find a way for delineating $S_0$ during the optimization process.
To remedy this issue, we leverage the rich 3D geometry information embedded in 3DToMolo, providing a fine-grained ability to precisely locate the substructure $S_0$. Formally, let $\textbf{X} = (\textbf{x}_1, \dots, \textbf{x}_n)$ denote the positions of the designated optimization sites. In this structural manipulation process, we employ a meticulous spatial translation approach. Around each $\textbf{x}_i$, a specified number of atoms are randomly initialized. The objective is to ensure that the center of mass (COM) of this newly initialized set of atoms matches the COM of the designated optimization sites:
$\bar{\textbf{X}} := \sum_i \textbf{x}_i /n,$
which is realized by Euclidean translation. Subsequently, a similar strategy is employed as in the previous section. The atoms situated specifically at the optimization sites and their associated chemical bonds are selectively noised, introducing controlled perturbations to the molecular structure. Meanwhile, all other aspects of the 2D and 3D structures within the molecule are held constant. This precise and localized perturbation strategy ensures that the optimization is targeted and confined solely to the designated sites while preserving the overall structural integrity of the molecule.

It is worth noting that we can even modify the atom numbers once we obtain $M_t$. For example, let $M_0$ denote the atom set of hydrogens and their related chemical bonds. Then, $M_t$ contains the noised atom set with random dropout or supplemental new atoms. During the denoising process, the 2D part inside the subgraph $S_0$ ($H_t$ and $E_t$) is adjusted, while the 3D positions are fine-tuned.

\paragraph{3D-Aware Substituent Group Optimization.}

A common template-based molecule synthesis pipeline involves two steps: 1. Finding the optimization site \cite{doi:10.1021/acscentsci.3c00572}; 2. Composing the selected fragments at the optimization site. This scenario imposes a highly nontrivial challenge on latent-space based representation methods for molecules \cite{liu2023moleculestm}, as locating the precise location of the optimization site while keeping the surrounding structure unchanged is difficult. However, leveraging \textbf{3D structure-based information} provides direct guidance for optimization by noising only the optimization area and generating the substituent group by moving random atoms to the optimization site. Formally, let $\textbf{v}_e$ denote the position of the optimization site. We initialize a set of atoms with 3D positions denoted by $\textbf{G}_T$ according to the stationary distribution of the noising process. We then translate $\textbf{G}_0$ to the optimization site position and start denoising:
\begin{equation}
\textbf{G}_T \rightarrow \textbf{G}_T + \textbf{v}_e \xrightarrow{df \text{guided Denoising}}  \textbf{G}_0^f.    
\end{equation}

The algorithms of 3DToMolo for three guidance optimization settings are provided in Section \ref{algorithms}.
\newpage

\section{Calculation of Redox Potentials}
\begin{figure}[H]
    \centering
    \includegraphics[width=0.9\linewidth]{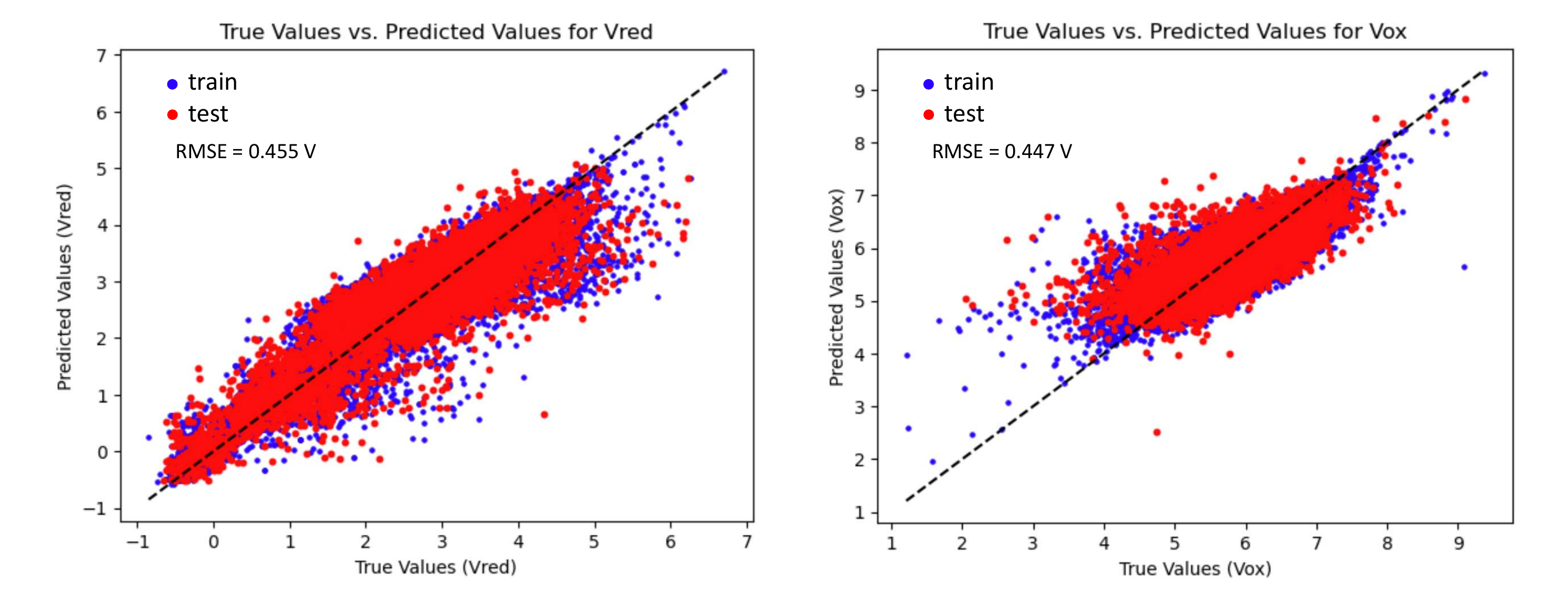}
    \caption{\label{fig:SI_redox predictor} Gradient boosting regression on the DFT calculated redox potentials. The dataset consists of 149 electrolyte additives for lithium-ion batteries \cite{redox-okamoto2018ab} and 41,801 molecules in the organic materials for energy application database (OMEAD) \cite{redox2-carvalho2023evolutionary}. Three quarters of the dataset are randomly sampled to be the train set while the rest are used as the test set. The root mean square errors (RMSE) on the test set are 0.455 V and 0.447 V respectively for the reduction potential (left) and the oxidation potential (right).}
\end{figure}

To rapidly assess the oxidation and reduction potentials of generated molecules, we employ machine learning techniques for qualitative structure-property relationship (QSPR) predictions based on straightforward structural descriptors. These descriptors, inspired by prior research \cite{redox-okamoto2018ab}, encompass factors such as the number of atoms of each element with the same coordination number, the count of rings with various sizes, the presence of aromatic rings, and the quantity of hydroxyl groups. A comprehensive list of descriptors is provided in Table \ref{tab:redox_descriptors}. The dataset used comprises DFT-calculated redox potentials, combining 149 electrolyte additives for lithium-ion batteries \cite{redox-okamoto2018ab} and 41,801 molecules from the Organic Materials for Energy Application Database (OMEAD) \cite{redox2-carvalho2023evolutionary}. We use the gradient boosting regression (GBR) method. The model is trained on three-quarters of the dataset and validated on the remaining quarter. In configuring the hyperparameters for the scikit-learn \footnote{https://scikit-learn.org/stable/} GBR module, we set the learning rate to 0.001, the number of estimators to 100,000, the maximum depth to 6, the minimum sample split to 3, and the maximum features to 2. On the test dataset, 3DToMolo yields a root mean square error (RMSE) of 0.455 V for the reduction potential and 0.447 V for the oxidation potential, as illustrated in Figure \ref{fig:SI_redox predictor}.

\begin{table}[H]
\begin{threeparttable}
\caption{Descriptors used in the redox potential regressor.}
    \centering
\label{tab:redox_descriptors} 
\begin{tabular}{ p{5cm} p{10.5cm}}
\hline\\[-4.8mm]\hline
\rowcolor{brown} Class & Descriptors \\
\hline
\rowcolor{brown!30} Atomic coordination number & B3, C4, C3, C2, N3, N2, N1, O2, O1, F1, Si4, P4, P3, S4, S3, S2, S1, Cl1 \\
\hline
\rowcolor{brown!15} Functional group \& substructure & Three- or four-member ring, five-member ring, six- or more-member ring, aromatic ring, hydroxyl group \\
\hline\\[-4.8mm]\hline
\end{tabular}
\begin{tablenotes}
      \small
      \item As in the referenced work \cite{redox-okamoto2018ab}, in the \textit{atomic coordination number} class, the letter stands for the element and the digit stands for the coordination number. For example, we count how many carbon atoms have a coordination number of 4 and use this number as the value for the descriptor "C4".
\end{tablenotes}
\end{threeparttable}
\end{table}

\newpage

\section{Similarity vs. denoising steps}
The similarity between two molecules is quantified by the Tanimoto coefficient based on the Morgan extended-connectivity fingerprints (ECFP) for the molecules \cite{bajusz2015tanimoto}. As references, given by Tanimoto coefficient (ECFP), the similarity between benzene and thiophene is $2/9\approx0.222$, and the similarity between naphalene and quinoxaline is 0.3125.

\begin{figure}[H]
    \centering
    \includegraphics[width=0.9\linewidth]{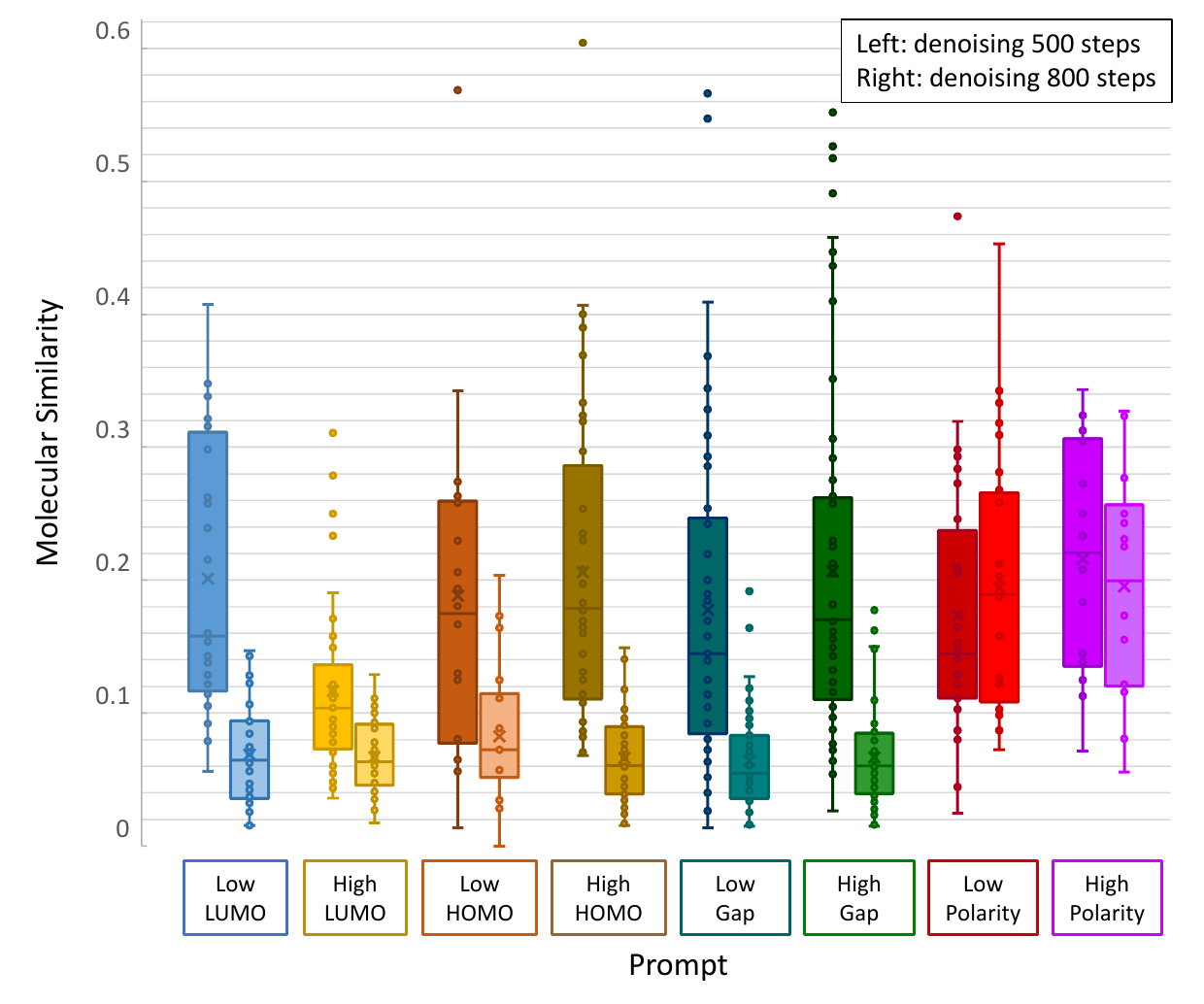}    \caption{\label{fig:SI_similarity_vs_denoising_steps} Box plots of molecular similarity between the optimized molecule and the original molecule in two settings: denoising 500 steps and denoising 800 steps. The average is marked by the symbol "$\times$" and the median is marked by the horizontal line. }
\end{figure}

\newpage

\section{More optimized results with non-coplanar benzene rings}

\begin{figure}[H]
    \centering
    \includegraphics[width=\linewidth]{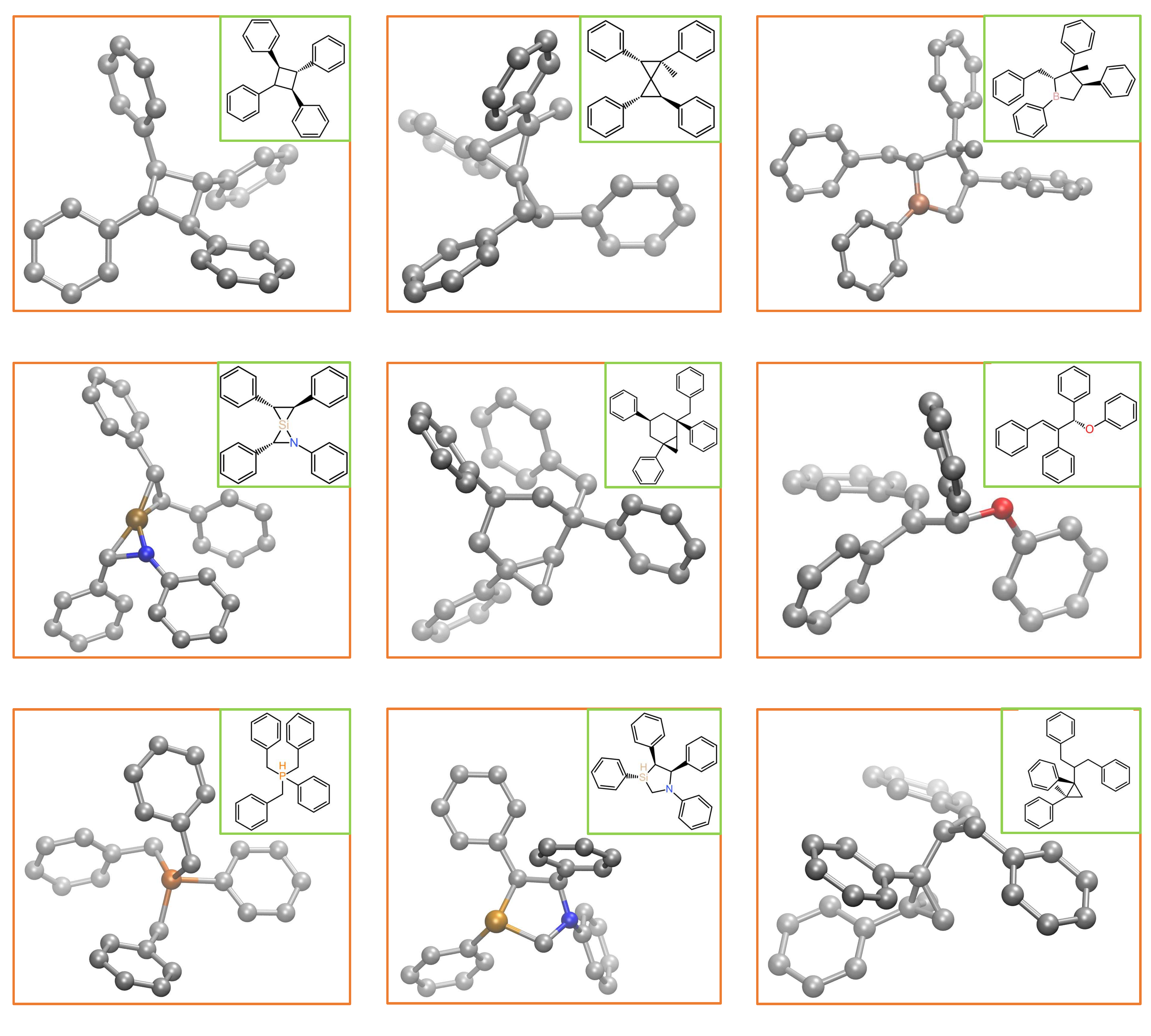}
    \caption{\label{fig:SI_more_examples_spiroconjugate} Selected optimized structures with non-coplanar benzene rings using the tetraphenylsilane molecule as the template.}
\end{figure}

\newpage

\section{Unconditional molecule optimization with structure constraints.}
\begin{figure}[H]
    \centering
    \includegraphics[width=\linewidth]{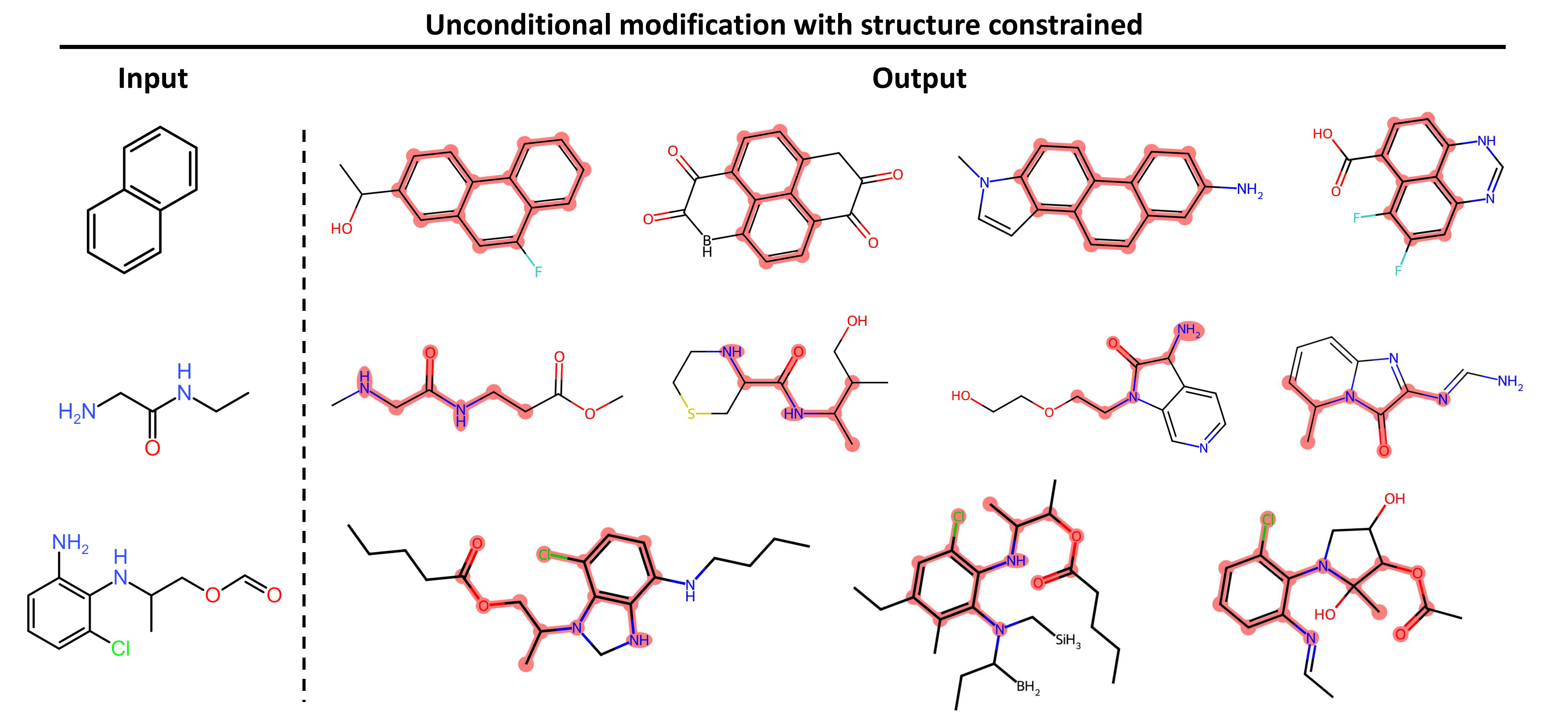}
    \caption{\label{fig:SI_unconditional} Selected optimized structures with non-coplanar benzene rings using the tetraphenylsilane molecule as the template.}
\end{figure}

We select three molecules for unconditional optimization while preserving their skeletons. As seen in Figure \ref{fig:SI_unconditional}, the output molecules exhibit a great variety in modifications, including generation of diverse heteroatoms, simultaneous substitutions on multi-sites, formation of cyclic structures, extension of aromatic rings and so on. Meanwhile, the seed skeletons are preserved successfully. We note that some of the complex outputs can hardly be synthesized from the input molecule. However, such unconditional optimization can be utilized as a tool to extensively search the chemical space for molecules that have the matching skeleton. The structure of the desired skeleton can be passed on to the model in the appearance of input molecule. In this sense, we should not be bothered much by the synthetic path from the input molecule to the optimization results.

\newpage

\section{Tetraphenylsilane optimization task using GPT3.5.}\label{sec:GPT_tetraphenylsilane}

\begin{table}[H]
\begin{threeparttable}
\centering
\caption{\label{tab:GPT_tetraphenylsilane} GPT3.5 optimization on the tetraphenylsilane molecule.}
\begin{tabular}{ p{11cm} c c}
\hline\\[-4.8mm]\hline
\rowcolor{brown} Prompt & Validity ratio & Hit ratio \\
\hline
\rowcolor{brown!30} 1) Please edit the central silicon atom of the tetraphenylsilane molecule while maintaining the non-coplanar structure of the four benzene rings. Please generate (another) 10 edited molecules in SMILES. & 0.64 & 0 \\
\hline
\rowcolor{brown!15} 2) Tetraphenylsilane is a molecule that has four non-coplanar benzene rings. Its SMILES is **smiles**. Please edit the central silicon atom of the tetraphenylsilane molecule while the edited molecules maintain four non-coplanar benzene rings. Please generate (another) 10 edited molecules in SMILES. 
 & 1.00 & 0 \\
\hline
\rowcolor{brown!30} 3) Please generate (another) 10 molecules that have four non-coplanar benzene rings in SMILES. & 1.00 & 0 \\
\hline
\rowcolor{brown!15} 4) Tetraphenylsilane is a molecule that has four non-coplanar benzene rings. Its SMILES is **smiles**. Please edit the internal region of the tetraphenylsilane molecule. The edited molecules must have four benzene rings that are non-coplanar. Please generate (another) 10 edited molecules in SMILES. & 1.00 & 0 \\
\hline
\rowcolor{brown!30} 5) Tetraphenylsilane is a molecule that has four non-coplanar benzene rings. Its SMILES is **smiles**. Please edit the internal region of the tetraphenylsilane molecule. The edited molecules must have four benzene rings. These four benzene must be non-coplanar. Please generate (another) 10 edited molecules in SMILES. & 1.00 & 0 \\
\hline
\rowcolor{brown!15} 6) Tetraphenylsilane is a molecule that has four non-coplanar benzene rings. Its SMILES is **smiles**. Please replace the central silicon atom by other substructure such that the edited molecules have four non-coplanar benzene rings. Please generate (another) 10 edited molecules in SMILES. & 1.00 & 0.70 \\
\hline
\rowcolor{brown!30} 7) Tetraphenylsilane is a molecule that has four non-coplanar benzene rings. Its SMILES is **smiles**. Please replace the central silicon atom by other substructure with more than one atom such that the edited molecules have four non-coplanar benzene rings. Please generate (another) 10 edited molecules in SMILES & 1.00 & 0 \\
\hline\\[-4.8mm]\hline
\end{tabular}
\begin{tablenotes}
      \small
      \item We asked GPT3.5 to generate 10 molecules everytime and repeated several times. From the second time and after, we added "another" to the prompt. In most cases, GPT3.5 fails because the generated molecule does not have four benzene rings or put them in linear connection. In those satisfactory cases, GPT3.5 is simply replacing the silicon atom by other atoms, including Ge, Sn, As, P, Pb, Bi and so on. The abbreviated SMILES string in the instructions is as follows:
      \item **smiles** = "c1ccc(c(c1)[Si](c2ccccc2)(c3ccccc3)c4ccccc4)".
\end{tablenotes}
\end{threeparttable}
\end{table}

\newpage
\section{Appointed-site optimization task using GPT3.5.}\label{sec:SI_GPTmultisite}

\begin{table}[H]
\begin{threeparttable}
\centering
\caption{\label{tab:GPT_penicillin} GPT3.5 editing on the penicillin molecule.}
\begin{tabular}{ p{11cm} c c}
\hline\\[-4.8mm]\hline
\rowcolor{brown} Prompt & Validity ratio & Hit ratio \\
\hline
\rowcolor{brown!30} 1) Please edit the penicillin molecule **smiles** without modifying the substructure **sub-smiles** so that it has a large electron withdrawing group, which may have low HOMO (Highest occupied molecular orbital) energy of the carbonyl group. Please generate (another) 10 edited smiles. & 0 & 0 \\
\hline
\rowcolor{brown!15} 2) Please edit the penicillin molecule **smiles** without modifying the substructure **sub-smiles** so that it has a large electron withdrawing group, which may have low HOMO (Highest occupied molecular orbital) energy of the carbonyl group. Please generate (another) 10 valid edited smiles. & 0 & 0 \\
\hline
\rowcolor{brown!30} 3) Please edit the penicillin molecule **smiles** so that it has a large electron withdrawing group next to the amide group, which may have low HOMO (Highest occupied molecular orbital) energy of the carbonyl group. Please generate (another) 10 valid edited smiles. & 0 & 0 \\
\hline
\rowcolor{brown!15} 4) Please edit the penicillin molecule **smiles** so that it has a better resistance to acids and lactamases. Please generate (another) 10 edited smiles. & 0 & 0 \\
\hline
\rowcolor{brown!30} 5) Please edit the penicillin molecule so that it has a better resistance to acids and lactamases. Please generate (another) 10 edited smiles. & 0 & 0 \\
\hline
\rowcolor{brown!15} 6) Please modify the benzyl group in the penicillin molecule so that the molecule has a large electron withdrawing group. Please generate (another) 10 edited molecules in SMILES. & 0 & 0 \\
\hline
\rowcolor{brown!30} 7) Please modify the benzyl group in the penicillin molecule while maintain the rest of the structure so that the molecule has a large electron withdrawing group. Please generate (another) 10 edited molecules in SMILES. & 0 & 0 \\
\hline
\rowcolor{brown!15} 8) Please modify the benzyl group in the penicillin molecule while maintain the rest of the structure so that the molecule has a large electron withdrawing group. An example of satisfactory editing is **example-smiles**. Please generate (another) 10 edited molecules in SMILES. & 0.20 & 0 \\
\hline
\rowcolor{brown!30} 9) Please modify the benzyl group in the penicillin molecule while maintain the rest of the structure so that the molecule has a large electron withdrawing group. An example of satisfactory editing is **example-smiles**. Please generate (another) 10 edited molecules in SMILES that you had never provided. & 0 & 0 \\
\hline\\[-4.8mm]\hline
\end{tabular}
\begin{tablenotes}
      \small
      \item We asked GPT3.5 to generate 10 molecules everytime and repeated several times. From the second time and after, we added "another" to the prompt. GPT3.5 failed to generate valid SMILES strings in most cases. None of valid cases has preserved the core structure and thus could not be regarded as satisfactory. The abbreviated SMILES strings in the instructions are as follows:
      \item **smiles** = "O=C(N[C@@H]1C(=O)N2[C@H](C(S[C@]12[H])(C)C)C(=O)O)CC1C=CC=CC=1".
      \item **sub-smiles** = "O=C(N[C@@H]1C(=O)N2[C@H](C(S[C@]12[H])(C)C)C(=O)O)".
      \item **example-smiles** = "CC1=C(C(=NO1)C2=CC=CC=C2)C(=O)N[C@H]3[C@@H]4N(C3=O)[C@H](C(S4)(C)C)C(=O)O".
\end{tablenotes}
\end{threeparttable}
\end{table}

\newpage
\section{Diversity of optimization results and multi-run performance}

\begin{figure}[H]
    \centering
    \includegraphics[width=\linewidth]{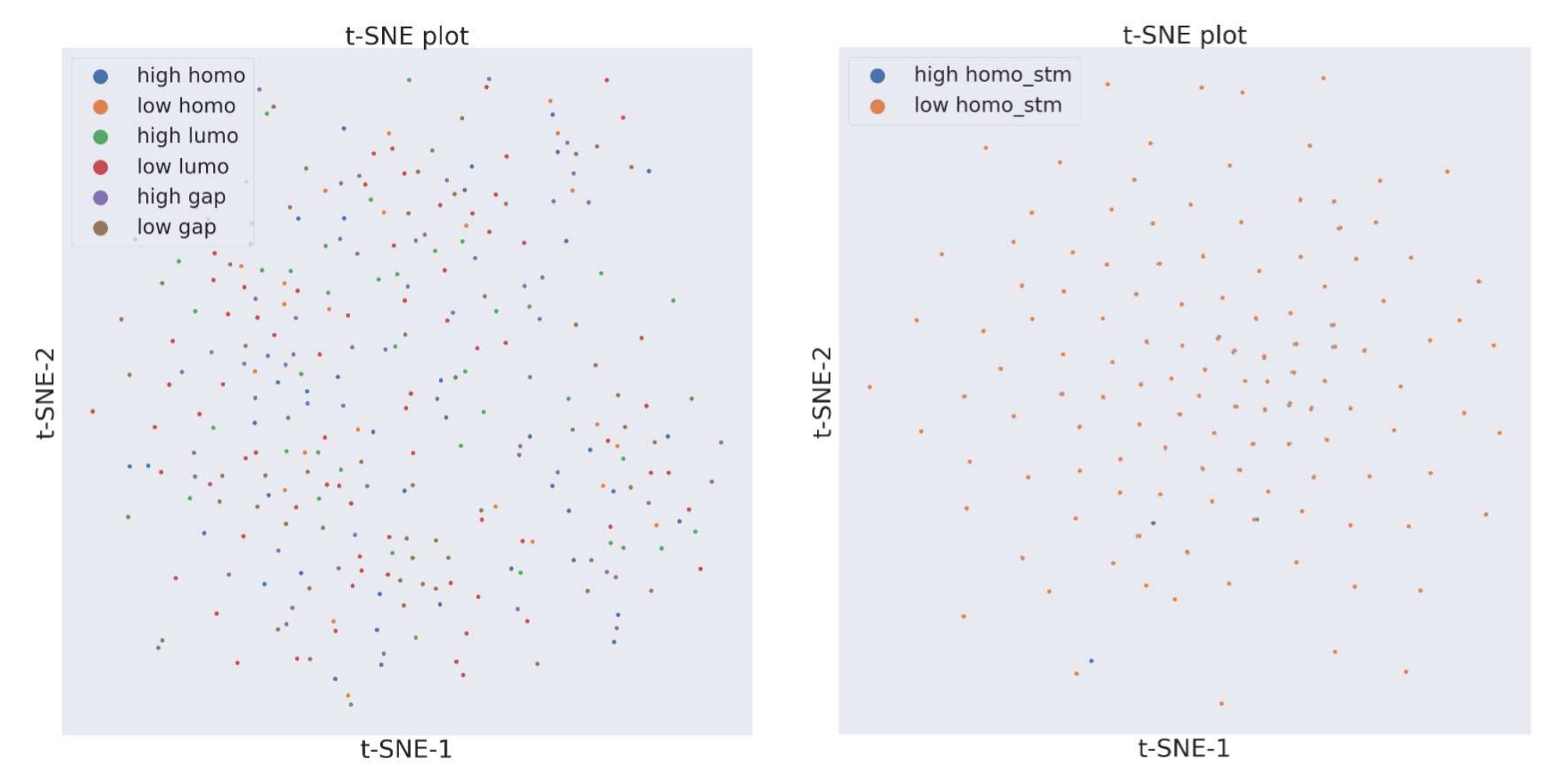}    \caption{\label{fig:SI_diversity} T-SNE plots of valid optimization results of 200 ZINC molecules under various prompts. In a t-SNE plot, each dot represents a molecule. The short distance between two dots means the corresponding molecules have similar structures. On the left is the plot for our 3DToMolo model. On the right is the plot for MoleculeSTM results that we reproduce. The overlaps of blue and orange dots in the right plot reflect that for most of input molecules, the output results by MoleculeSTM are in fact the same, even though the prompts are opposite.}
\end{figure}

\begin{table}[H]
  \caption{Multi-run performance evaluation compared with Single-run}
    \centering
  \label{tab:multirun}
  \begin{tabular}{ p{8cm} c c c}
    \hline\\[-4.8mm]\hline
    \rowcolor{brown} Prompts & MoleculeSTM & GPT3.5 & 3DToMolo \\
    \cline{1-4}
    \rowcolor{brown!30} This molecule has low HOMO (Highest occupied molecular orbital) value, which is more stable. & 35.00 -> 35.00 & 37.50 -> 72.00 & 46.50 -> 75.00\\
    \cline{1-4}
    \rowcolor{brown!15} This molecule has high HOMO (Highest occupied molecular orbital) value, which is more reactive and susceptible to electron acceptance or participation in chemical reactions. & 28.00 -> 28.00 & 40.50 -> 82.50 & 58.50 -> 97.50 \\
    \cline{1-4}
    \rowcolor{brown!30} This molecule has more hydrogen bond donors.  & 05.00 -> 05.00 & 12.50 -> 74.00 & 34.50 -> 78.00\\
    \cline{1-4}
    \rowcolor{brown!15} This molecule has high permeability and has high HOMO-LUMO gap value, which is insulating or non-conductive. The large energy difference between the HOMO and LUMO orbitals makes it less likely for electrons to be excited across the gap, resulting in low electrical conductivity.  & 01.00 -> 01.00 & 18.00 -> 44.00 & 33.50 -> 94.00 \\
    \hline\\[-4.8mm]\hline
    \end{tabular}
\end{table}

\newpage
\section{Summary of multi-objective molecule optimization results}
\begin{table}[!htbp]
  \caption{Multi-objective molecule optimization}
    \centering
  \label{tab:multiobj}
  \begin{tabular}{ p{11cm} c c c}
    \hline\\[-4.8mm]\hline
    \rowcolor{brown} Prompts & MoleculeSTM & GPT3.5 & 3DToMolo \\
    \cline{1-4}
    \rowcolor{brown!30} This molecule is soluble in water and has high polarity. & 08.00 & 68.00 & 31.50\\
    \cline{1-4}
    \rowcolor{brown!15} This molecule is insoluble in water and has low polarity. & 06.00 & 59.00 & 95.00 \\
    \cline{1-4}
    \rowcolor{brown!30} This molecule has high permeability and has high HOMO-LUMO gap value, which is insulating or non-conductive. The large energy difference between the HOMO and LUMO orbitals makes it less likely for electrons to be excited across the gap, resulting in low electrical conductivity.  & 01.00 & 44.00 & 94.00 \\
    \cline{1-4}
    \rowcolor{brown!15} This molecule is soluble in water and has more hydrogen bond acceptors. & 01.00 & 66.50 & 55.00 \\
    \cline{1-4}
    \rowcolor{brown!30} This molecule is insoluble in water and has more hydrogen bond acceptors. & 01.00 & 24.00 & 15.00 \\
    \hline\\[-4.8mm]\hline
    \end{tabular}

\end{table}

\newpage
\section{Synthetic accessibility of optimization results}

\begin{figure}[!htbp]
    \centering
    \begin{minipage}[c]{\textwidth}
    \centering
    \includegraphics[width=0.75\textwidth]{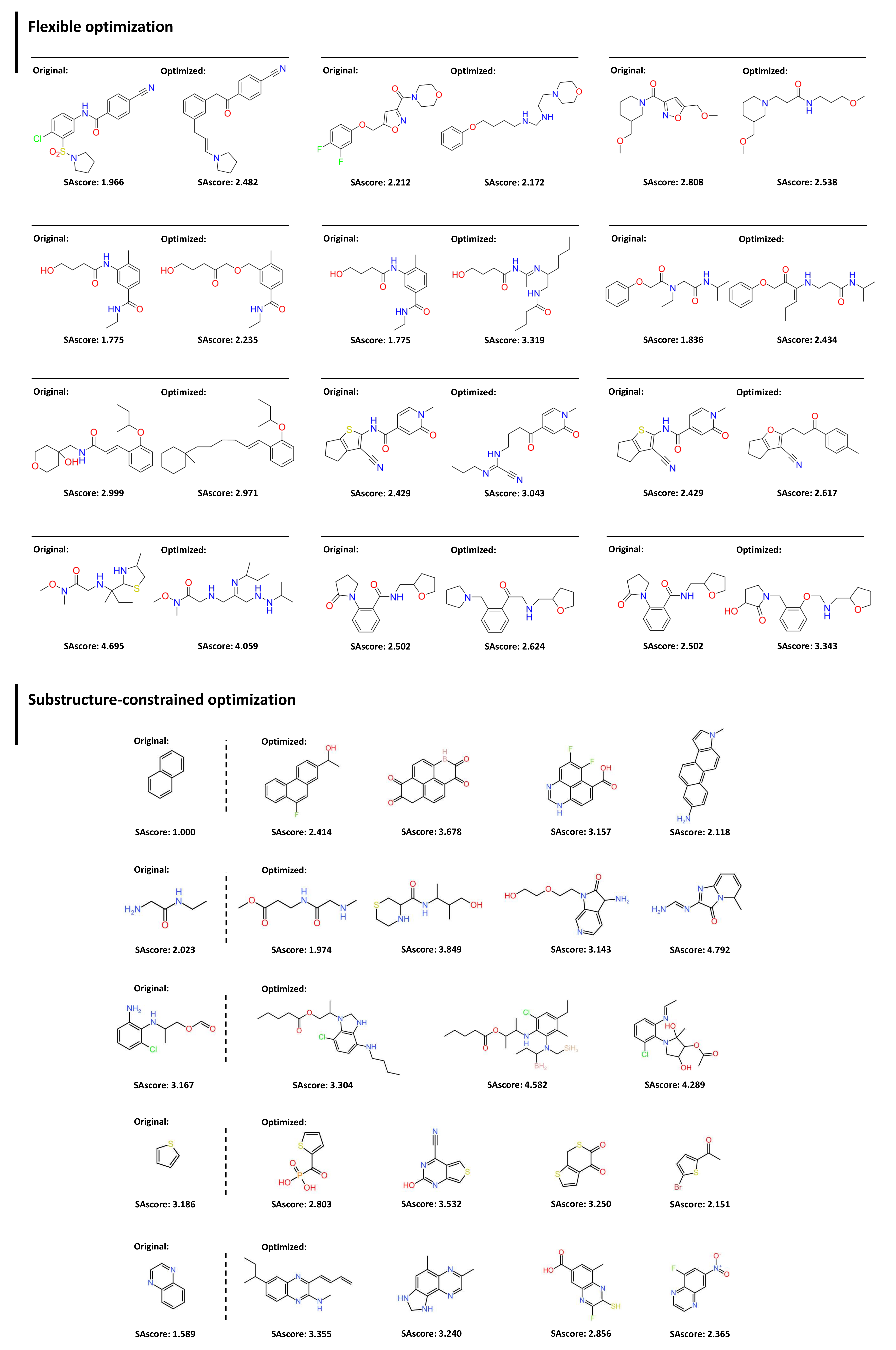}
    \end{minipage}
    \caption{A summary of the synthetic accessibility score (SAscore) for presented molecules in the main article. The SAscore ranges from 1 to 10, where 1 means "easy to make" and 10 means "very difficult to make". As references, SAscores for natural products mainly range from $5\sim6$ and SAscores for bioactive molecules mainly range from $3\sim4$. \cite{SAscore}}
    \label{fig:SI_SAscore} 
\end{figure}

\newpage
\section{Validity of our pretrained generative model}

\begin{table}[!htbp]
\setlength{\tabcolsep}{5pt}
\fontsize{9}{9}\selectfont
\centering
\vspace{-2ex}
\caption{\small{Validation results on PCQM4Mv2.}}
\label{tab: pretrain_val}
 \resizebox{\columnwidth}{!}{
\begin{tabular}{l c c c c c c c c c}
\toprule
Model & Mol stable $\uparrow$ & Atom stable $\uparrow$ & Validity $\uparrow$ & Unique $\uparrow$ & AtomTV $\downarrow$ & BondTV $\downarrow$ & ValW1 $\downarrow$ & Bond Lengths W1 $\downarrow$ & Bond Angles W1 $\downarrow$ \\
\midrule
EDM & 55.0 & 92.9 & 34.8 & 100.0 & 0.212 & 0.049 & 0.112 & 0.002 & 6.23 \\
3DToMolo \cite{hoogeboom2022equivariant} & \textbf{84.5} & \textbf{99.6} & \textbf{80.0} & \textbf{100.0} & \textbf{0.059} & \textbf{0.021} & \textbf{0.008} & \textbf{0.003} & \textbf{2.16} \\
\bottomrule
\end{tabular}}
\end{table}
We provide standard metrics for testing the performance of 3DToMolo's diffusion model. The molecule's stability and validity are measured by default Rdkit \cite{Landrum2020} algorithms. We also provide $W_1$ and total variation distances between our generated molecules' 2D topology and the training set, proposed by the referenced work \cite{vignac2023midi}. The hyper-parameters for training the baseline EDM \cite{hoogeboom2022equivariant} follow the same setting as the referenced work \cite{vignac2023midi}. As we can see from the table, 3DToMolo achieves significantly better performance than EDM, which demonstrates the effectiveness of our proposed equivariant transformer for generating valid and diverse molecules in the chemical space.

\newpage
\section{Validation of backbone model on chirality distinction}

\newpage
\section{Experimental Details}

\textbf{Hyperparameters} We list the key hyperparameters used for 3DToMolo pretraining with 3D molecular graphs and text as inputs, which is presented in Table \ref{tab: Hyperparameters}.

\begin{table}[!htbp]
\setlength{\tabcolsep}{5pt}
\fontsize{9}{9}\selectfont
\centering
\vspace{-2ex}
\caption{\small{Hyperparameters of pretraining.}}
\label{tab: Hyperparameters}
 \resizebox{0.6\columnwidth}{!}{
\begin{tabular}{c c}
\toprule
Hyperparameters & Value \\
\midrule
epochs & $32$ \\
learning rate for LlaMa & $1e-5$ \\
learning rate for chemical structure branch & $1e-5$ \\
loss function & InfoNCE \\
diffusion\_steps & $800$ \\
llm\_hidden\_size & $4096$ \\
\bottomrule
\end{tabular}}
\end{table}

\newpage
\section{Algorithms of 3DToMolo for three guidance optimization settings}
\label{algorithms}

\begin{algorithm}
\renewcommand{\algorithmicrequire}{\textbf{Input: }}
\renewcommand{\algorithmicensure}{\textbf{Output:}}
\caption{Flexible molecule optimization under text prompts}
\label{Algorithm 1}
\begin{algorithmic}[1]

        \Require Given a molecule to be optimized $M_0:=(H_0, E_0, P_0)$  and text-prompt $y$ implies the orientation of optimization.   
        \State Sampling atoms to input: $H_0, E_0, P_0 \leftarrow H_0 \cup H_{ex}, E_0 \cup E_{ex}, P_0 \cup P_{ex}$
        \For {$t:=1$ to $T$}
            \State $z_{t-1} \leftarrow (E_{t-1}, H_{t-1})$
            \State $H_t, E_t \leftarrow \mathcal{C}(z_{t-1}Q_t)$
            \State $P_t \leftarrow \sqrt{\bar{\alpha_t}} P_0 + \sqrt{1 - \bar{\alpha_t}} \epsilon$  \Comment{Forward process}
        \EndFor

        \State $M_T^{\prime} \leftarrow M_T$
        \For {$t:=T$ to $1$}    \Comment{Conditional denoising process}
            \State $q_{\theta}(M_{t-1}^{\prime}| M_t^{\prime}, y) \propto q_{\theta}(M_{t-1}^{\prime}| M_t^{\prime}) \cdot e^{-\lambda <\nabla_{M_t^{\prime}} ||y - f(M_t^{\prime})||^2  , M_{t-1}^{\prime}>}$  
            \State $M_{t-1}^{\prime} \sim q_{\theta}(M_{t-1}^{\prime}| M_t^{\prime}, y)$ \Comment{Sample based on the differential of CLIP loss}
        \EndFor
        \Ensure Output optimized molecule  $M_0^{\prime}$ with desired property $y$.
        
\end{algorithmic}
\end{algorithm}

\begin{algorithm}
\renewcommand{\algorithmicrequire}{\textbf{Input: }}
\renewcommand{\algorithmicensure}{\textbf{Output:}}
\caption{Molecule optimization with Structural constraints}
\label{Algorithm 2}
\begin{algorithmic}[1]

        \Require Given a molecule to be optimized $M_0^c:=(H_0^c, E_0^c, P_0^c)$ with substructure $S_0$ to be protected  and text-prompt $y$ implies the orientation of optimization.  
        
        \State $M_0^c \leftarrow M_0 \cup S_0, M_0:=(H_0, E_0, P_0)$
        \State Sampling atoms to $M_0$: $H_0, E_0, P_0 \leftarrow H_0 \cup H_{ex}, E_0 \cup E_{ex}, P_0 \cup P_{ex}$
        
        \For {$t:=1$ to $T$}
            \State $z_{t-1} \leftarrow (E_{t-1}, H_{t-1})$
            \State $H_t, E_t \leftarrow \mathcal{C}(z_{t-1}Q_t)$ 
            \State $P_t \leftarrow \sqrt{\bar{\alpha_t}} P_0 + \sqrt{1 - \bar{\alpha_t}} \epsilon$  \Comment{Forward process}
        \EndFor

        \State $M_T := (H_T \cup S_0, E_T \cup S_0, P_T \cup S_0)$  \Comment{Substructure remain fixed}

        \State $M_T^{\prime} \leftarrow M_T$
        \For {$t:=T$ to $1$}    \Comment{Conditional denoising process}
            \State $q_{\theta}(M_{t-1}^{\prime}| M_t^{\prime}, y) \propto q_{\theta}(M_{t-1}^{\prime}| M_t^{\prime}) \cdot e^{-\lambda <\nabla_{M_t^{\prime}} ||y - f(M_t^{\prime})||^2  , M_{t-1}^{\prime}>}$  
            \State $M_{t-1}^{\prime} \sim q_{\theta}(M_{t-1}^{\prime}| M_t^{\prime}, y)$ \Comment{Sample based on the differential of CLIP loss}
            \State $M_{t-1}^{\prime}[S_0] \leftarrow M_T[S_0]$
        \EndFor
        \Ensure Output optimized molecule  $M_0^{\prime}$ with desired property $y$.
        
\end{algorithmic}
\end{algorithm}

\begin{algorithm}
\renewcommand{\algorithmicrequire}{\textbf{Input: }}
\renewcommand{\algorithmicensure}{\textbf{Output:}}
\caption{Hard-coded molecule optimization on appointed sites}
\label{Algorithm 3}
\begin{algorithmic}[1]

        \Require Given a molecule to be optimized $M_0^c:=(H_0^c, E_0^c, P_0^c)$ with editing site $X$ and text-prompt $y$.     
        \State Extract a subgraph of input surrounding $X$: $M_0^s \leftarrow Subgraph(M_0^c| X, k)$ \Comment{K-hop subgraph}
        \State $M_0 \leftarrow M_0^s[X], M_0:=(H_0, E_0, P_0)=(H_0^s[X], E_0^s[X], E_0^s[X])$
        \State Sampling atoms to $M_0$: $H_0, E_0, P_0 \leftarrow H_0 \cup H_{ex}, E_0 \cup E_{ex}, P_0 \cup P_{ex}$
        \State Match the center of mass (COM) of $(H_{ex}, E_{ex}, P_{ex})$ with $\bar{\textbf{X}} := \sum_i \textbf{x}_i /n$ 
        
        \For {$t:=1$ to $T$}
            \State $z_{t-1} \leftarrow (E_{t-1}, H_{t-1})$
            \State $H_t, E_t \leftarrow \mathcal{C}(z_{t-1}Q_t)$ 
            \State $P_t \leftarrow \sqrt{\bar{\alpha_t}} P_0 + \sqrt{1 - \bar{\alpha_t}} \epsilon$  \Comment{Forward process}
        \EndFor
        \State $M_T := (H_T \cup M_0^s, E_T \cup M_0^s, P_T \cup M_0^s)$ 

        \State $M_T^{\prime} \leftarrow M_T$
        \For {$t:=T$ to $1$}    \Comment{Conditional denoising process}
            \State $q_{\theta}(M_{t-1}^{\prime}| M_t^{\prime}, y) \propto q_{\theta}(M_{t-1}^{\prime}| M_t^{\prime}) \cdot e^{-\lambda <\nabla_{M_t^{\prime}} ||y - f(M_t^{\prime})||^2  , M_{t-1}^{\prime}>}$  
            \State $M_{t-1}^{\prime} \sim q_{\theta}(M_{t-1}^{\prime}| M_t^{\prime}, y)$ \Comment{Sample based on the differential of CLIP loss}
            \State $M_{t-1}^{\prime}[M_0^s \backslash X] \leftarrow M_T[M_0^s \backslash X]$
        \EndFor
        \Ensure Output optimized molecule  $M_0^{\prime}$ with desired property $y$.
        
\end{algorithmic}
\end{algorithm}

\newpage
\bibliography{SI_references}